\documentclass{article}

\usepackage{arxiv}

\usepackage[utf8]{inputenc} 
\usepackage[T1]{fontenc}    

\usepackage[colorlinks=true, allcolors=blue]{hyperref}
\newcommand{\email}[1]{\href{mailto:#1}{\tt{\nolinkurl{#1}}}}
\newcommand{\orcid}[1]{ORCID: \href{https://orcid.org/#1}{\tt{\nolinkurl{#1}}}}

\usepackage{url}            
\usepackage{booktabs}       
\usepackage{amsfonts}       
\usepackage{nicefrac}       
\usepackage{microtype}      
\usepackage{lipsum}
\usepackage{amssymb}
\usepackage{latexsym}

\usepackage{url}
\usepackage{xcolor}
\usepackage{hyperref}

\usepackage{amsmath}
\usepackage{graphicx}
\usepackage{cleveref}
\usepackage[export]{adjustbox}
\usepackage{subfig}
\usepackage{mwe}
\usepackage{graphicx}
\usepackage{pdflscape}

\title{Wise-SrNet: A Novel Architecture for Enhancing Image Classification by Learning Spatial Resolution of feature maps}

\author{
 Mohammad Rahimzadeh \\
  Entropy Mind Co\\
  \texttt{\email{mr7495@yahoo.com}} \\
  \texttt{\orcid{0000-0002-8550-8967}} \\
  \texttt{Corresponding author}
  \And
 Soroush Parvin \\
  School of Electrical Engineering\\
  Iran University of Science and Technology, Iran\\
  \texttt{\email{soroush.parvin.qu@gmail.com}}
  \And
 Amirali Askari \\
  School of Electrical Engineering\\
  Iran University of Science and Technology, Iran\\
  \texttt{\email{amiraliaskari2014@gmail.com}}
  \And
 Elnaz Safi \\
  Entropy Mind Co\\
  \texttt{\email{e_safi@entropymindai.com}} \\
  \texttt{\orcid{0000-0002-4786-5234}} 
  \And
 Mohammad Reza Mohammadi \\
  School of Computer Engineering\\
  Iran University of Science and Technology, Iran\\
  \texttt{\email{mrmohammadi@iust.ac.ir}}\\
  \texttt{\orcid{0000-0002-1016-9243}} 
}
  
\begin{document}
\maketitle
\begin{abstract}
One of the main challenges since the advancement of convolutional neural networks is how to connect the extracted feature map to the final classification layer.
VGG models used two sets of fully connected layers for the classification part of their architectures, which significantly increased the number of models' weights.
ResNet and the next deep convolutional models used the Global Average Pooling (GAP) layer to compress the feature map and feed it to the classification layer. Although using the GAP layer reduces the computational cost, but also causes losing spatial resolution of the feature map, which results in decreasing learning efficiency. 
In this paper, we aim to tackle this problem by replacing the GAP layer with a new architecture called Wise-SrNet. It is inspired by the depthwise convolutional idea and is designed for processing spatial resolution while not increasing computational cost.
 We have evaluated our method using three different datasets:  Intel Image Classification Challenge, MIT Indoors Scenes, and a part of the ImageNet dataset. We investigated the implementation of our architecture on several models of the Inception, ResNet, and DenseNet families. Applying our architecture has revealed a significant effect on increasing convergence speed and accuracy.
 Our Experiments on images with 224×224 resolution increased the Top-1 accuracy between 2\% to 8\% on different datasets and models.
Running our models on 512×512 resolution images of the MIT Indoors Scenes dataset showed a notable result of improving the Top-1 accuracy within 3\% to 26\%. We will also demonstrate the GAP layer's disadvantage when the input images are large and the number of classes is not few. In this circumstance, our proposed architecture can do a great help in enhancing classification results. 
The code is shared at \href{https://github.com/mr7495/image-classification-spatial}{https://github.com/mr7495/image-classification-spatial}.
\end{abstract}



\section{Introduction}

Convolutional layers have made a considerable improvement in computer vision tasks. Before introducing convolutional layers, deep learning was known for using fully connected (FC) layers. One of the main problems of FC layers was utilizing too many parameters (weights), which made the developers unable to extend the number of layers due to the possibility of overfitting and also the need for huge RAM storage.

Convolutional layers' most important novelty was decreasing the number of weights while improving learning efficiency, so the developers could design models with more layers that achieved more reliable results on different benchmarks. 
LeNet\cite{726791} and AlexNet\cite{krizhevsky2012imagenet} were the leaders for introducing convolutional models in Image classification tasks. After them, VGG \cite{simonyan2015deep} designed an architecture made by convolutional layers and obtained a good improvement in image classification on the ImageNet benchmark\cite{5206848}. 

VGG models generate a 7×7×512 feature map from an input image with 224×244×3 resolution. VGG is made of almost 14 million weights for the feature extraction base. At the final layers, VGG architectures flatten the feature map to a 25088 neuron tensor and use two FC layers, each with 4096 neurons, to connect the flattened feature map to the final classification layer. The biggest issue in this conversion is increasing the number of weights from 14 million to almost 138 million! This increment for just three layers is not optimized at all.

At the next level, ResNet models \cite{he2015deep}, inspired by the architecture of VGG, introduced new residual layers and designed a new series of convolution models.
These models generate a 7×7×2048 feature map from an input image with 224×224×3 resolution. Based on the feature map size, the authors couldn't use FC layers at the end of their models because the number of models' parameters would have become very high, and the performance would be turned awful.
 Another innovation of the ResNet models was using the Global Average Pooling (GAP) layer instead of FC layers at the final layers.

GAP was previously created by \cite{lin2013network}, which proposed this idea to average between the spatial values of each channel of the extracted feature map to compress the features.
For a tensor array with a shape equal to 7×7×2048, we call the 7×7 windows and 2048, the kernels and channels, respectively. So this tensor is made of channel data and spatial data. Spatial data is related to each of the 7×7=49 elements of each channel, while the channel data is related to each spatial element in different channels (between 1 to 2048).

Utilizing the GAP layer on ResNet models allowed the authors to extend the number of channels and also reduce computational cost. However, this approach was efficient but caused the model to lose spatial data due to the averaging between each channel's spatial resolution.
After ResNet, other proposed architectures in the future like DenseNet \cite{huang2018densely}, NasNet \cite{zoph2018learning}, Inception \cite{szegedy2016rethinking, szegedy2016inceptionv4, chollet2017xception}, MobileNet \cite{howard2017mobilenets}, and EfficientNet \cite{tan2020efficientnet} families also applied the GAP layer at the end of their models.

Another revolution in deep learning was the introduction of separable convolutional layers \cite{chollet2017xception}.
Separable convolutional layer proposes the idea that each filter of a convolutional layer doesn't have to be applied to both spatial and channel data simultaneously. It is constructed of a depthwise convolutional layer and a pointwise convolutional layer. 
This approach did almost the same mutation to convolutional layers as convolutional layers did to FC layers because of remarkably decreasing the number of weights.

Here, we mention an example for better understanding. Consider applying a convolution layer with 512 filters and 3×3 kernel to a tensor array with  20×20×256 shape to get an output array with shape 20×20×512. If we use a simple convolutional layer, then for each filter of the convolution layer, 3×3×256 weights should be trained, and in total, the number of weights would become 3×3×256×512=1179648 weights. 

Instead, if we use a separable convolutional layer, it applies a 3×3 kernel of weights for each of the 256 channels separately.  A pointwise convolutional layer will be then applied to the depthwise convolution layer's output to achieve a tensor array with a shape of 20×20×512. So, the total number of weights will be equal to 3×3×256 (for the depthwise convolutional layer) + 256×512 (for the pointwise convolutional layer) =133376 weights. Note we did not count the bias parameters.

By comparing 1179648 weights with 133376 weights, it will be clear that using separable convolutional layers will significantly reduce the number of model weights. From the date these layers were created, some models like Xception \cite{chollet2017xception}, and NasNet \cite{zoph2018learning} used them and achieved remarkable results. 

In this work, we want to address the problem of losing spatial data (caused by using GAP layers) without increasing computational cost. Our goal is to make the models able to learn how to combine the extracted feature map data and feed it to the final classification layer without losing any information. Inspired by the architecture of depthwise convolution layers, we have designed a series of layers that enables the models to learn how to analyze both of the spatial and channel information of the extracted feature map. We have called our novel architecture Wise-SrNet.

We replaced the GAP layer with a depthwise convolutional layer with specific configurations. Our developed depthwise convolutional layer gives the model the ability to learn some weights for compressing the final feature map without losing spatial information.
 In this while, we faced that the model may become overfitted due to the large kernel size of the depthwise convolutional layer.
For removing this issue, we also added some constraints and other layers to the depthwise convolutional layer, and these layers together considerably improved the classification accuracy while not increasing computational cost. Our proposed techniques can be applied to all the deep convolutional models, and so we can create more robust models and much better benchmark results on the ImageNet or other challenges. 
We tested our methods with several models from three famous convolutional families of ResNet, DenseNet, and Inception. As the results show improvement in accuracy, and efficiency, our method can be integrated with various computer vision tasks other than image classification, that require feature compression such as ensemble learning \cite{rahimzadeh2021roct, sobti2021ensemv3x}, scene recognition \cite{seong2020fosnet}, object detection, etc.

 In our experiments, we observed that using the GAP layer in datasets with many classes and large images is not reliable and may fail in many circumstances.  In these conditions, our proposed architecture can do a great help for image classification tasks.

These papers \cite{qiu2018global,Card_2019}, proposed Global Weighted Average Pooling (GWAP) layers instead of GAP. In these researches, the GWAP weights will be indicated based on the values of the previous layers. The main difference between these works and ours is that the weighted-averaging layers' weights are equal for each channel of the feature map. It means the same weighted averaging matrix will be applied to each channel of the feature map. In contrast, in our architecture, different weights for each channel will be trained regarding the fact that the model must learn how to contribute the data of each channel independently in producing output array. Another difference is that our work is not based on performing averaging on the final feature map data, and we let the model learn how to process the feature map when training. The function the model creates for processing the feature map data can be anything and is not limited to a single function like averaging.

Another research \cite{peeples2021histogram} proposed a histogram-based architecture to replace the GAP layer and improve classification accuracy. Their architecture is made of two main convolution layers, an average pooling layer alongside some other layer for extracting the histogram information from the final feature map. Then this histogram feature map will be fed to the final classification layer.
Their work's main problem is that the two convolutional layers increase the number of weights, especially when the number of classes rises, computational cost increases notably.

Our paper is organized as follows: In section \ref{2}, we will describe the neural networks and their limits, and our proposed architectures, and section \ref{3}, represents the utilized datasets and our experimental results. In section \ref{4}, the paper is discussed, and in section \ref{5}, we have concluded our paper.

\section{Materials and Methods}
\label{2}

In this section, we first describe the usually used architectures for compressing the final feature maps (\ref{221}, \ref{222}), and then in \ref{223}, we talk about the depthwise separable convolutional layer that is the core idea of our method and its advantage. In subsection \ref{224}, the idea for preventing the models from overfitting has been proposed, and section \ref{225} explains the computational costs of different architectures integrated with our proposed method and compares it with previous methods.

\subsection{Global Average Pooling}
\label{221}

\begin{figure}[!ht]
\centering
\subfloat[Classification with Global Average Pooling]{\label{gap}\includegraphics[width=\linewidth]{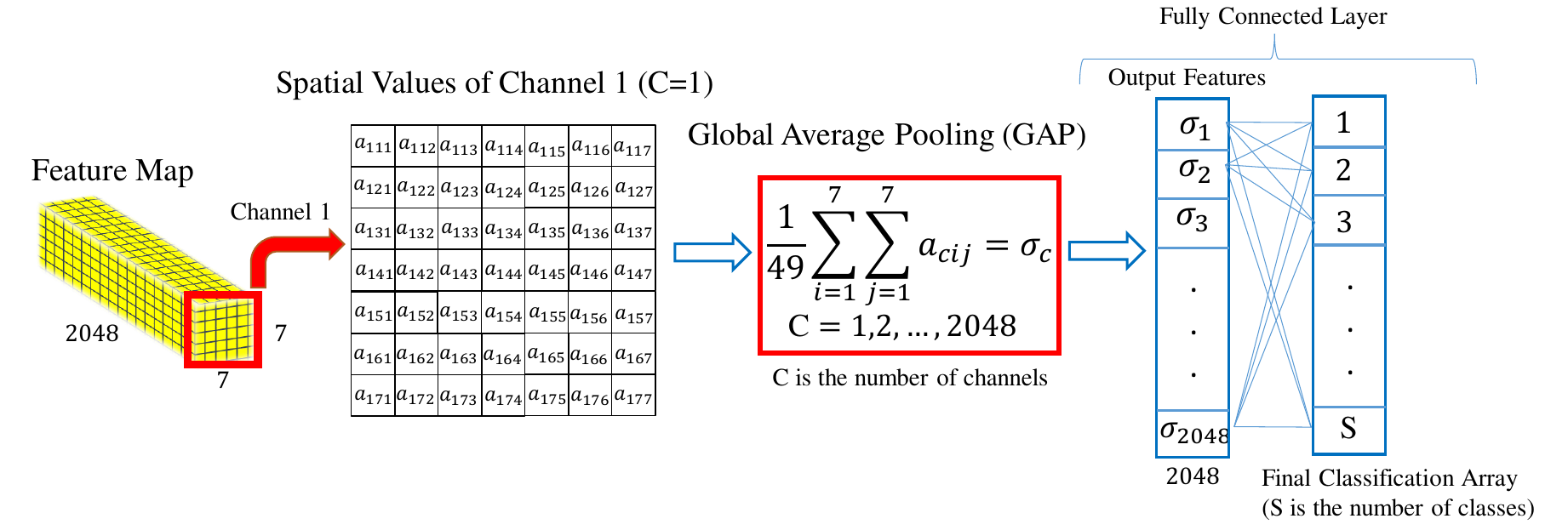}}
\newline
\subfloat[Classification with Global Weighted Average Pooling]{\label{gwap}\includegraphics[width=\linewidth]{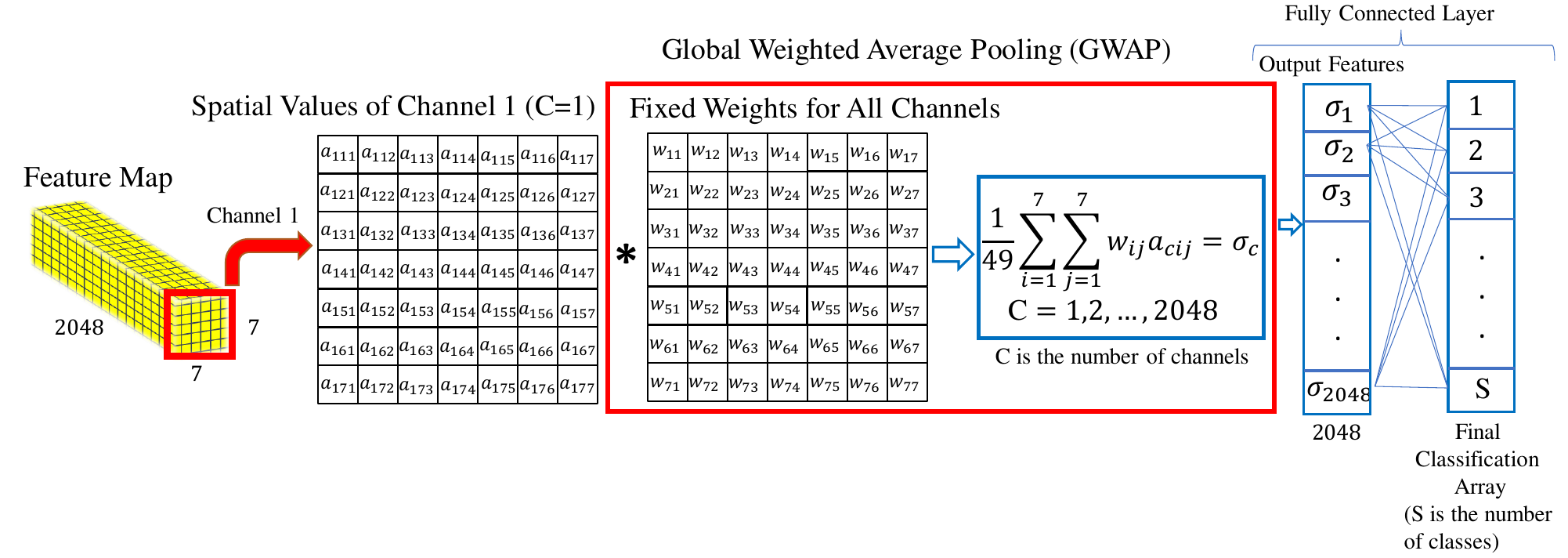}}
\newline
\subfloat[Extracting spatial data with depthwise convolutional layer]{\label{depthwise}\includegraphics[width=\linewidth]{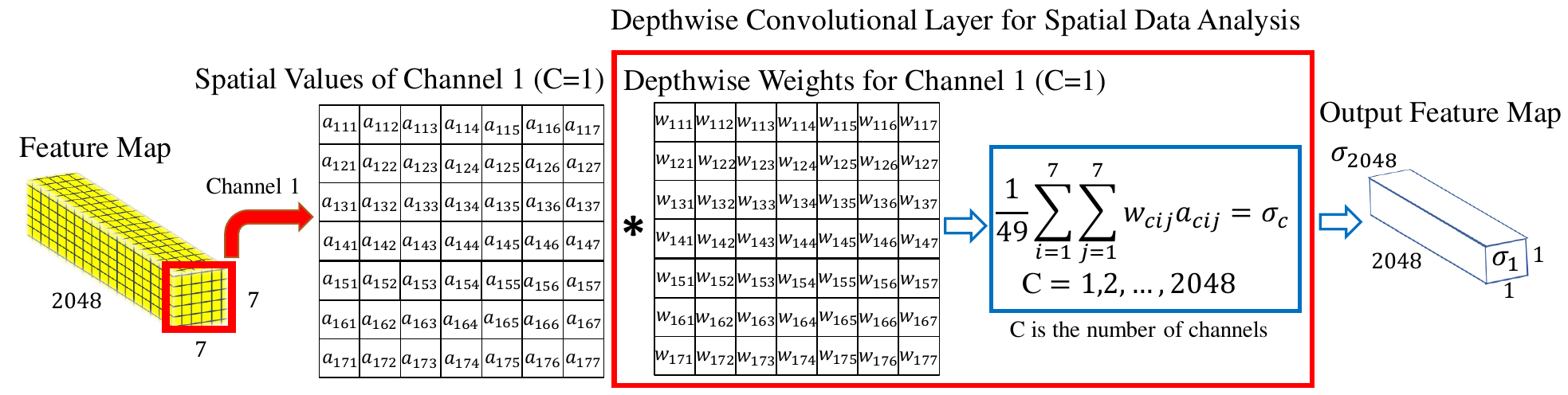}}

\caption{The architectures of different classification layers with GAP, GWAP, and depthwise convolutional layer are portrayed in this figure.}
\label{3base}
\end{figure}

Global Average Pooling (GAP) \cite{lin2013network} proposes a technique to average between the feature map's spatial values for compressing the size of feature maps. After VGG models, most of the proposed deep convolutional classifiers implemented this layer at the end of their architecture. Although this layer decreases the computation cost, it removes a part of the image information, resulting in reaching lower accuracy. The architecture of Global Average Pooling (GAP) is depicted in figure \ref{gap}.

The noteworthy point is that how much the kernel size of the feature maps is more extensive, the loss of spatial information will be more. For example, in a 7×7×2048 feature map, the GAP layer calculates the mean of 7×7=49 values for each channel. In another case, for a 16×16×2048 feature map, the GAP layer averages between 16×16=256 values of each channel. Averaging within 256 values will remove much more information than averaging between 49 values. This means the weakness of GAP layers will be more significant when the input images, and so the feature maps are larger.

\subsection{Flattening and Fully Connected Layers}
\label{222}

\begin{figure}[!ht]
\centering
\includegraphics[width=\linewidth]{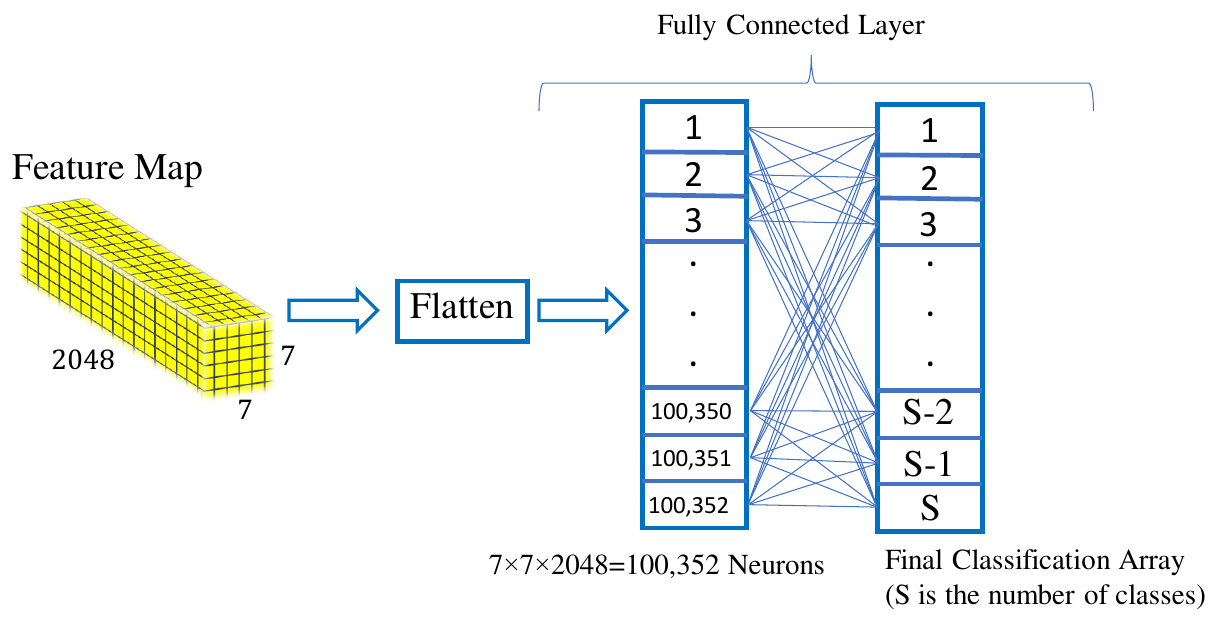}
\caption{Classification with Flatten and Fully-Connected layers}
\label{dense}
\end{figure}

Many other researchers preferred using fully connected layers (FC Layers) instead of Global Average Pooling (GAP) layers. To use these layers, first, they flatten the extracted feature map to a one-dimensional array. Then they will use an FC layer to connect the flattened feature map to the final classification layer  (figure \ref{dense}). This way, the model can see all the features and will not lose any data. figure \ref{dense}, presents the architecture of this method.

But what is wrong with this technique? This technique is only appliable when there are few classes and input images are not large. Because when we have more classes, using this technique makes the number of weights much higher. Consider having a feature map with 7×7×2048 resolution (extracted from a 224×224 image) and 100 classes of images. The flattened feature map will have 7×7×2048=100,352 neurons, and the number weights for the final FC layer will be: 7×7×2048×100=10,035,200. It means that the number of total weights will be increased by more than 10 million neurons.

If the number of classes would be even higher, e.g., the ImageNet dataset includes 1000 classes, then the number of
weights will be nearly 100 million more! Also, if the input image resolution is higher than 224 (which in many cases can be higher), the output feature map will be even larger, and so the number weights will be more.

This increment will weaken the model because of two reasons: 
\begin{enumerate}
    \item We will need much more RAM space to train and
test the models, and the training process will be much more time-consuming.
    \item Having more weights for just one layer will result in the model’s overfitting or underfitting and make the model
unable to learn, causing the output results to be even worth compared to the GAP layer.
\end{enumerate}

\subsection{Depthwise Convolutional Layer for Spatial Resolution Analysis}
\label{223}

In this research, we aim to explore a solution to help the classifiers do not lose the feature map's spatial data while keeping almost the same computational cost.  In other words, we wish to propose an idea that is like the GAP layer in computational cost but is more accurate by processing the spatial values of the feature map wisely.

The best way to produce this idea is to create a layer that allows the deep classifiers to learn how to combine and weighted-sum the final feature map's data and compress it to a smaller array. The GAP layer performance is fixed, and it only calculates the average of the feature map spatial values (e.g., GAP converts a 7×7×2048 array to a one-dimension array with 2048 values, meaning it averages between the 49 spatial resolution values of each channel).
Suppose the model could learn how to allocate some weights to the feature map's spatial resolution while training to find the best way to process and compress the feature map. In that case, the model first processes the feature map's spatial data and then processes the channel data in the next layer (the layer that connects it to the final classification layer). Figures \ref{depthwise}, and \ref{224-depthwise} illustrate this idea. 
 
We present the idea of using a specific depthwise convolutional layer after getting the final feature map, which gives the model this ability to learn how to work with the feature map's spatial data without losing information. What makes our proposed architecture different from the Global Weighted Average Pooling layer (GWAP) \cite{qiu2018global}, is that the GWAP layer allocates some fixed weights to the spatial data of all the channels (figure \ref{gwap}). It means in a 7×7×2048 feature map, 49 weights could be learned for processing the feature map spatial values, and these weights were fixed for all the 2048 channels. Applying the same weights to all the channels makes the model deficient because, channels are different from each other, and some may be more important than others. In this situation, the weights that will be applied to the spatial values of different channels should vary from each other, and the model must learn these weights while training.

Our proposed architecture solves this problem by making the model capable of learning different weights for each channel separately. In a 7×7×2048 feature map, the model learns 49 different weights for each channel (figure \ref{depthwise}). In other words, the model investigates each of the 2048 channels independently and learns 49 weights for processing the spatial data of each channel differently. Using this architecture helps the classifier explore all the channels' spatial data alongside the channels data to make the best prediction. It is noteworthy that our architecture's spatial process is not based on averaging at all. The model learns how to combine and sum these data, and it may like any operations and is not limited to a single function like averaging. figure \ref{depthwise} illuminates the mentioned issues.

For implementing our proposed idea, in the first stage, after getting the feature map, we feed it to a depthwise convolutional layer with a kernel size equal to the feature map's spatial shape. For example, if the feature map's shape is 7×7×2048, we fed it to a depthwise convolutional layer with a kernel size equal to 7×7. In this way, this layer will apply a 7×7 set of learnable weights for each channel to process each channel's data separately. The output of this layer will be a 1×1×2048 tensor array.

 After that, we apply a flatten layer to reduce the output tensor's dimensions (convert the 1×1×2048 array to an array with 2048 neurons). Then we will connect the one-dimensional layer to the final classification layer, which is a fully connected layer and contain neurons equal to the number of classes (e.g., in a 70 classes dataset, the final layer is a FC layer with 70 neurons). figure \ref{224-depthwise}, shows the explained architecture.

\subsection{Constraints for Overfitting Reduction}
\label{224}

After executing several investigations, we noticed that although using the expressed architecture in \ref{223} led to higher accuracy; the models faced overfitting after passing several training epochs. Based on our experiments, the overfitting mainly comes from the large kernel size of the depthwise convolutional layers. Besides, if we wish to train the models on larger images like 512×512 resolution, the generated feature map's kernel size and the depthwise convolution layer's kernel size would become 16×16, so the overfitting will be even more. Here we come with two solutions:

\begin{enumerate}

\item Applying a constraint to the depthwise convolutional layer so that its weights do not become negative while training. As the depthwise convolution layer's weights are applied to the feature map neurons, it is better that the weights do not become negative because the summation of negative weights with positive weights can waive some information. This constraint helps a lot in diminishing model overfitting. 

\item Placing an Average pooling layer before the depthwise convolutional layer. It makes the kernel size of the feature map and the next depthwise convolutional layer smaller. Reducing the depthwise convolutional layer's kernel size to 3×3, 4×4, or 5×5 will decrease the overfitting possibility, so the whole model learning quality will be more reliable. 
For 224×224 images, we set the average pooling kernel to 2×2 to change the feature map's and the depthwise convolutional layer's kernel size from 7×7 to 3×3. On 512×512 images, as the feature map's spatial size will be 16×16, we changed the average pooling layer kernel to 3×3 in order to make the output feature map's spatial size, 5×5.

\end{enumerate}

Researchers can also test other values for choosing the average pooling layer's kernel size, which may lead to better performance. Our goal in this paper is to show that implementing these techniques improves learning effectiveness.
It must be noted that based on our experiments, using bias parameters in the depthwise convolutional layers and not using activation functions will obtain better results. No regularization algorithm was also utilized. The final proposed architecture will be like figure \ref{224-avg-depthwise-const}.

\subsection{Computational Costs}
\label{225}

In table. \ref{cost}, we explore and compare the computational costs of our proposed architectures for Xception \cite{chollet2017xception}, ResNet50 \cite{he2015deep} and DenseNet121 \cite{huang2018densely} models. The numbers in this table are based on having 224×244 resolution input images and 70 classes.

From this Table, we can see that models with our architecture contain almost the same number of parameters and FLOPs as models with the GAP layer. Xception with our architecture has 23,751,622 trainable parameters, while Xception with the GAP layer contains 23,731,142 trainable parameters, which are almost equal.
This means that our architecture is capable of processing all the feature map's data, and not losing spatial information without increasing computational cost.

Based on the table. \ref{cost}, ResNet50, DenseNet121, and Xception models with our architecture have 0.08\%, 0.14\%, and 0.09\% more weights than the same models with GAP, respectively. Increasing model weights by less than 0.2\% will not affect the computational costs.

\begin{table*}[!ht]
\centering
\large
\caption{This table shows the number of parameters (weights) / FLOPs of three models using different classification layers. These numbers are based on having 70 classes of images with 224×224 resolution.
 Base model refers to the feature extraction part without applying any classification layer. Base model with Global Average Pooling means applying the GAP in the classification layers for features compression. Base Model with Depthwise Convolution is the model with the depthwise convolutional layer for analyzing spatial data. 
Base Model with Average Pooling and Depthwise Convolution expresses the usage of a pre-averaging layer with 2×2 kernel size for diminishing the feature map spatial size and preventing the depthwise convolution layer from overfitting.}
\begin{adjustbox}{width=1\linewidth}
\begin{tabular}{|l|l|l|l|l|}
\hline
\begin{tabular}[c]{@{}l@{}}Model\\ Name\end{tabular} & \begin{tabular}[c]{@{}l@{}}Feature\\ Extraction\\ Base Model\end{tabular} & \begin{tabular}[c]{@{}l@{}}Classification Model\\               with\\ Global Average Pooling\end{tabular} & \begin{tabular}[c]{@{}l@{}}Classification Model \\                       with\\ Depthwise Separable Convolution\end{tabular} & \begin{tabular}[c]{@{}l@{}}Classification Model \\       with Average Pooling and\\ Depthwise Separable Convolution\end{tabular} \\ \hline
ResNet50                                             & 23,587,712                                                                & 23,731,142 / 7.75 G                                                                                        & 23,833,542 / 7.75 G                                                                                                          & 23,751,622 / 7.75G                                                                                                               \\ \hline
Xception                                             & 20,861,480                                                                & 21,004,910 / 9.13 G                                                                                        & 21,107,310 / 9.13 G                                                                                                          & 21,025,390 / 9.13 G                                                                                                              \\ \hline
DenseNet121                                          & 7,037,504                                                                 & 7,109,254 / 5.70 G                                                                                         & 7,160,454 / 5.70 G                                                                                                           & 7,119,494 / 5.70 G                                                                                                               \\ \hline
\end{tabular}
\end{adjustbox}

\label{cost}
\end{table*}

\section{Experimental Results}
\label{3}

\subsection{Dataset}
\label{21}

We have used three datasets for our investigations:

\begin{enumerate}
  \item A part of the ImageNet dataset \cite{5206848} \footnote{ This dataset is shared at \url{https://www.kaggle.com/mohammadrahimzadeh/imagenet-70classes}}
  \item Intel image classification challenge \cite{intel} \footnote{ This dataset is shared at \url{https://www.kaggle.com/puneet6060/intel-image-classification}}
  \item MIT Indoors Scenes \cite{quattoni2009recognizing}  \footnote{ This dataset is shared at \url{https://www.kaggle.com/itsahmad/indoor-scenes-cvpr-2019}}
\end{enumerate}

The details of these datasets have been described in Table. \ref{dataset}.

We utilized three different datasets to investigate various criteria and prove that our architecture will enhance classification results generally. 

The ImageNet dataset \cite{5206848} is a popular and large dataset that contains lots of images. It is usually the reference dataset for evaluating deep convolutional models. The developers who train their models on this dataset have access to several powerful GPUs; otherwise, the training procedure may be impossible and very time-consuming due to the existence of around 1.3 million images in this dataset. As we did not have these facilities, we selected 70 of 1000 classes of this dataset for our experiments. The ImageNet dataset carries between 1000-1500 training images and 50 validation images for each class, but our selected dataset contains 500 training images and 50 validation images for each class and 31500 training images, and 3500 validation images in total. However, our training images are less than the ImageNet dataset, but the number of validation images for each class is the same.

Although we explored our models on two other datasets, as the ImageNet dataset is a reference dataset in image classification tasks, we also wished to evaluate our architecture on some part of this dataset to show our proposed methods' ultimate ability.

The Intel Image Classification \cite{intel} is another dataset we used for our investigations. Intel corporation provided this dataset to create another benchmark in image classification tasks. It contains 14034 training images and 3000 validation images belonging to six classes which are: buildings, forest, glacier, mountain, sea, and street. The default images of this dataset are in  150×150 pixels resolution. We utilized the whole dataset for training and validation procedures.

MIT university has also introduced a new dataset for image classification benchmarks named MIT Indoors Scenes \cite{quattoni2009recognizing}. MIT Indoors Scenes includes 67 categories of images from different scenes and views like bookstore, garage, gym, library, restaurant, office, etc. There are 5360 training images and 1340 validation images in this dataset. The main challenge of this dataset is the existence of few images (80 images) per class compared to other datasets. This situation makes classification difficult and the learning process weak. We have also investigated our models on this dataset to show our architecture's performance under challenging circumstances.

\begin{table*}
\centering
\large
\caption{Details of the utilized datasets are presented in this table.}
\begin{tabular}{|l|l|l|l|}
\hline
Dataset                              & Number of  Classes & Training Images & Test Images \\ \hline
Sub-ImageNet                         & 70                 & 31500           & 3500           \\ \hline
Intel Image Classification Challenge & 6                  & 14034           & 3000           \\ \hline
MIT Indoors Scenes                  & 67                 & 5360            & 1340           \\ \hline
\end{tabular}
\label{dataset}
\end{table*}

We implemented our models on \href{https://colab.research.google.com/}{Google Colaboratory Notebooks}, which allocated a Tesla P100 GPU, 2.00GHz Intel Xeon CPU, and 12GB RAM on Ubuntu server to us. We used  Keras library \cite{chollet2015keras} on Tensorflow backend \cite{tensorflow2015-whitepaper} for developing and running models.

Data augmentation was used for all the datasets to enhance the training procedure. All the modes used the same data augmentation techniques as detailed in Table. \ref{data augment}. To show our architecture's ultimate performance, we selected various models from the three great families of convolutional neural networks: ResNet, DenseNet, and Inception families for our experiments. We wish to prove that our proposed ideas are capable of improving every deep convolutional network.

\begin{table*}
\centering
\large
\caption{This table shows the data augmentation techniques that we utilized for training the models.}
\begin{tabular}{|l|l|}
\hline
Technique         & Range / Usage \\ \hline
Horizontal Flip   & Yes           \\ \hline
Vertical Flip     & Yes           \\ \hline
Zoom              & 20 \%         \\ \hline
Rotation          & 360 degree    \\ \hline
Width Shift       & 10 \%         \\ \hline
Height Shift      & 10\%          \\ \hline
Channel Shift     & 50 pixels     \\ \hline
Brightness Change & 0 - 1.2       \\ \hline
Preprocessing     & Yes (Caffe)   \\ \hline
\end{tabular}

\label{data augment}
\end{table*}

We have divided our experiments into two categories based on the resolution of the input images: 224×224 and 512×512.

\subsection{Experiments on Images with 224×224 Pixels Resolution}
\label{31}

In the ImageNet image classification benchmark, it is usually common to resize the images to 224×224 resolution to run the models. 
Although some models like InceptionV3 \cite{szegedy2016rethinking} converted ImageNet images to a different size (299×299), most of the proposed deep convolutional networks chose 224 as the images side size.

Based on this fact, in this stage of our work, all the three datasets images were resized to 224×244 resolution. 
 The details about all the parameters used for each dataset are listed in Table. \ref{param}.

\begin{table*}[!hb]
\centering
\large
\caption{This table shows the parameters adopted for training the models on 224x224 images.}
\begin{adjustbox}{width=1\linewidth}
\begin{tabular}{l|l|l|l|}
\cline{2-4}
                                                                                                                      & Sub-ImageNet                                                                              & MIT Indoors Scenes                                                                         & Intel Image Classification                                                                \\ \hline
\multicolumn{1}{|l|}{Batch Size}                                                                                      & 70                                                                                        & 70                                                                                        & 70                                                                                        \\ \hline
\multicolumn{1}{|l|}{Image Size}                                                                                      & 224x224                                                                                   & 224x224                                                                                   & 224x224                                                                                   \\ \hline
\multicolumn{1}{|l|}{Optimizer}                                                                                       & SGD (momentum:0.9)                                                                        & SGD (momentum:0.9)                                                                        & SGD (momentum:0.9)                                                                        \\ \hline
\multicolumn{1}{|l|}{Initial Learning rate}                                                                           & 0.045                                                                                     & 0.045                                                                                     & 0.045                                                                                     \\ \hline
\multicolumn{1}{|l|}{Learning rate decay}                                                                             & 0.94 every 2 epochs                                                                       & 0.94 every 2 epochs                                                                       & 0.94 every 2 epochs                                                                       \\ \hline
\multicolumn{1}{|l|}{Epochs}                                                                                          & 148                                                                                       & 200                                                                                       & 220                                                                                       \\ \hline
\multicolumn{1}{|l|}{Transfer Learning}                                                                               & No                                                                                        & No                                                                                        & No                                                                                        \\ \hline
\multicolumn{1}{|l|}{\begin{tabular}[c]{@{}l@{}}Base Model\\ Weight Initializer\end{tabular}}                         & \begin{tabular}[c]{@{}l@{}}Keras \\ Random Initialization\end{tabular}                    & \begin{tabular}[c]{@{}l@{}}Keras \\ Random Initialization\end{tabular}                    & \begin{tabular}[c]{@{}l@{}}Keras \\ Random Initialization\end{tabular}                    \\ \hline
\multicolumn{1}{|l|}{\begin{tabular}[c]{@{}l@{}}Depthwise Convolution Layer\\ Kernel Weight Initializer\end{tabular}} & \begin{tabular}[c]{@{}l@{}}Random Normal Initialization\\ (Mean:0, Std:0.01)\end{tabular} & \begin{tabular}[c]{@{}l@{}}Random Normal Initialization\\ (Mean:0, Std:0.01)\end{tabular} & \begin{tabular}[c]{@{}l@{}}Random Normal Initialization\\ (Mean:0, Std:0.01)\end{tabular} \\ \hline
\multicolumn{1}{|l|}{\begin{tabular}[c]{@{}l@{}}Depthwise Convolution Layer\\ Bias Initializer\end{tabular}}          & Zero Bias Initializtion                                                                   & Zero Bias Initializtion                                                                   & Zero Bias Initializtion                                                                   \\ \hline
\end{tabular}
\end{adjustbox}

\label{param}
\end{table*}

As described in Table. \ref{param}, we set the batch size to 70, and we used SGD optimizer with a momentum of 0.9. We selected an adaptive learning rate inspired by \cite{chollet2015keras} with an initialize value of 0.045 and a decay of 0.94 per every two epochs. 

The important point is that we did not use pre-trained weights at the beginning of the training procedure because the available pre-trained weights are from the trained models on the ImageNet dataset and images with 224×224 resolution using the GAP layer. These weights are trained based on the GAP layer for many epochs meaning the model's whole weights are trained somehow to be suitable for global averaging at the end.
As we want to compare the performance of our architecture with the normal condition (using the GAP layer), the comparison will not be fair when the initialized wights are produced and converged utilizing the GAP layer.

We wish to prove that training the models with our architecture from scratch will result in better learning and classification accuracy, which can create a novel development in image classification benchmarks. Therefore all the weights were initialized using Random initialization functions. 

\begin{figure}[!ht]
\centering
\subfloat[Deep convolution model with the GAP layer at its classification section.]{\label{224-gap}\includegraphics[width=\linewidth]{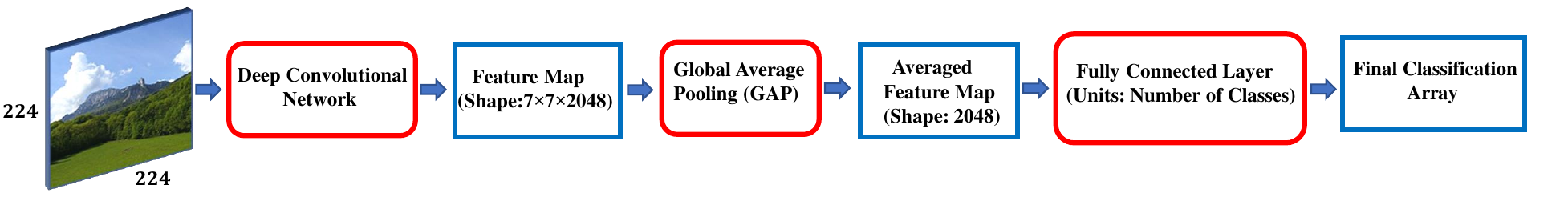}}
\newline
\subfloat[Deep convolution model with the GAP and dropout layers at its classification section.]{\label{224-gap-dp}\includegraphics[width=\linewidth]{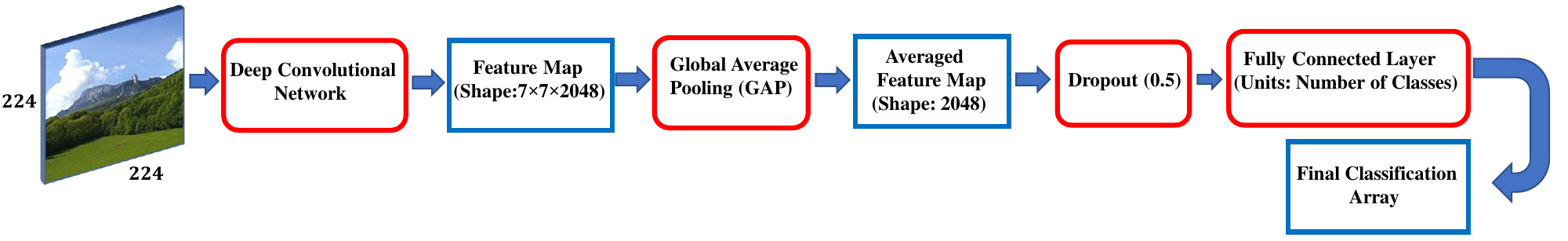}}
\newline
\subfloat[Deep convolution model with the proposed depthwise convolutional layer for fixing spatial data loss problem.]{\label{224-depthwise}\includegraphics[width=\linewidth]{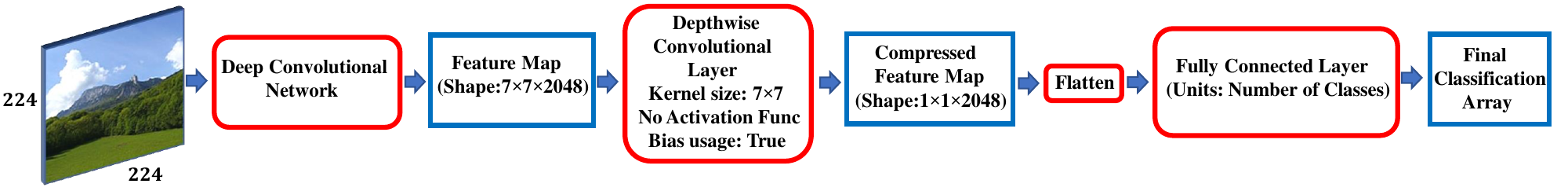}}
\newline
\subfloat[Deep convolution model with the proposed depthwise convolutional layer for extracting feature map's spatial and channel data and a non-negative constraint to prevent overfitting.]{\label{224-depthwise-const}\includegraphics[width=\linewidth]{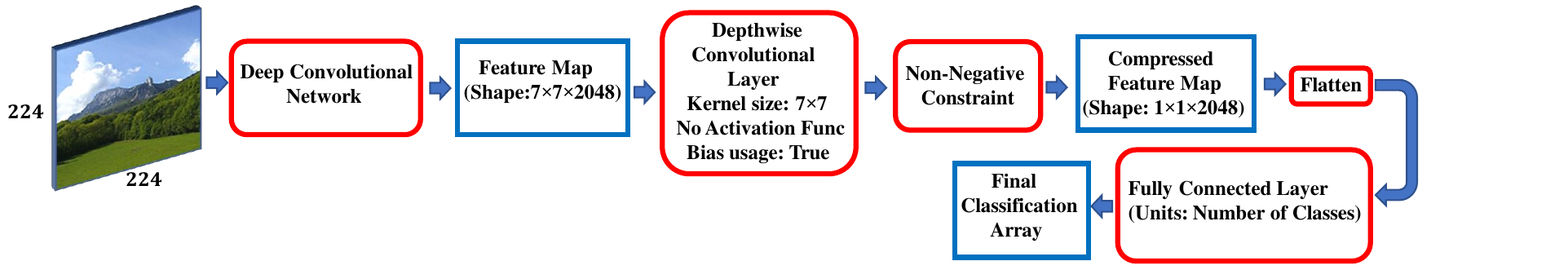}}
\newline
\subfloat[Deep convolution model with the proposed depthwise convolutional layer for extracting feature map's spatial and channel data and a non-negative constraint and pre-averaging layer to prevent overfitting.]{\label{224-avg-depthwise-const}\includegraphics[width=\linewidth]{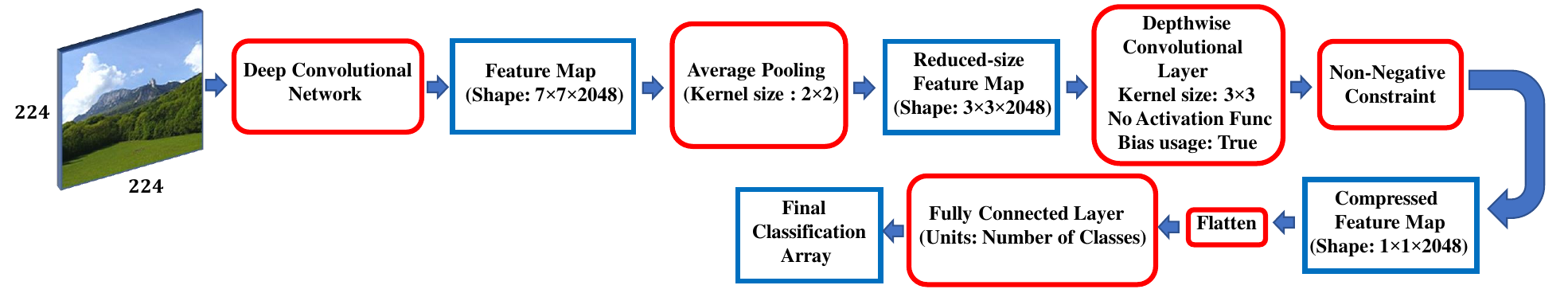}}
\newline
\subfloat[Deep convolution model with the proposed depthwise convolutional layer for extracting feature map's spatial and channel data. This model is enhanced by non-negative constraint, pre-averaging, and dropout layers.]{\label{224-avg-depthwise-const-dp}\includegraphics[width=\linewidth]{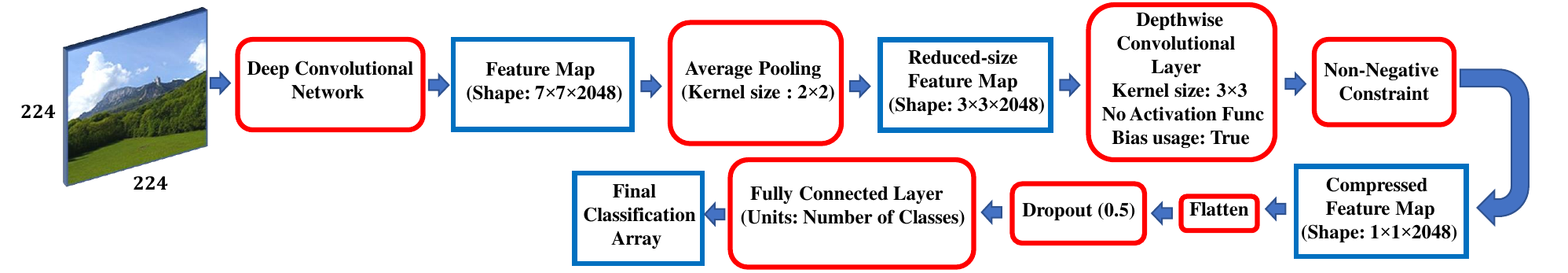}}

\caption{These figures represent the sex architecture we implemented for our experiments on images with 224×224 resolution. The red boxes are the model computational layers, and the blue boxes show the output arrays.}
\label{224-all}
\end{figure}

In this section, we investigate six versions of classifiers for each model:

\begin{enumerate}
  \item Base model with GAP layer: This is the simple classification method of using the GAP layer for compressing the features and feeding them to the final classification layer. The architecture is depicted in figure \ref{224-gap}. \label{v1}
  \item Base model with GAP layer and 50\% Dropout:
This architecture adds a dropout layer to the Base model with GAP layer. figure \ref{224-gap-dp} presents this architecture. \label{v2}
  \item Base model with Depthwise Convolutional layer: This architecture shows the usage of our proposed depthwise convolutional layer for extracting spatial data and feeding the enhanced and compressed feature map to the final FC layer. figure \ref{224-depthwise} clarifies the idea. \label{v3}
  \item Base model with Depthwise Convolutional layer and Non-negative constraint: This architecture is related to applying the non-negative constraint to the depthwise convolutional layer, which forces its weights not to become negative while training. This technique was implemented to improve learning and decrease overfitting. figure \ref{224-depthwise-const} represents the architecture. \label{v4}
  \item Base model with Averaging, Depthwise Convolutional layer, and Non-negative constraint:
 In this architecture, a pre-averaging layer and non-negative constraint are applied to the depthwise convolutional layer for enhancing learning effectiveness and prevent overfitting. figure \ref{224-avg-depthwise-const} expresses the architecture. \label{v5}
  \item Base model with Averaging, Depthwise Convolutional layer, Non-negative constraint, and 50\% Dropout: This the same as the last architecture with this difference of utilizing a dropout layer. (figure \ref{224-avg-depthwise-const-dp}) \label{v6}
  \item Base model with GWAP: This is another architecture trying to solve the same problem. \label{v7}
\end{enumerate}

In the following, we will describe the results of each dataset separately. For each dataset, three models of the popular deep convolutional networks of DenseNet, ResNet, and Inception families were chosen for training and validating our architectures.

\subsubsection{Sub-ImageNet Dataset}
\label{311}

We selected 70 classes of the ImageNet dataset, including 31500 training images and 3500 images, for validating the models' performance. This selected dataset can be accessed at \href{https://www.kaggle.com/mohammadrahimzadeh/imagenet-70classes}{https://www.kaggle.com/mohammadrahimzadeh/imagenet-70classes}.
We trained the models for 148 epochs on this dataset.

Three models (DenseNet121 \cite{huang2018densely}, ResNet50 \cite{he2015deep} and Xception \cite{chollet2017xception})  were selected for experimenting on this dataset. For each of these three model, we evaluated seven different versions:

\begin{itemize}
  \item Base model with GAP layer (version \ref{v1}, figure \ref{224-gap}) 
  \item Base model with GAP layer and 50\% Dropout (version \ref{v2}, figure \ref{224-gap-dp}) 
  \item Base model with Depthwise Convolutional layer (version \ref{v3}, figure \ref{224-depthwise})
  \item Base model with Depthwise Convolutional layer and Non-negative constraint (version \ref{v4}, figure \ref{224-depthwise-const})
  \item Base model with Averaging, Depthwise Convolutional layer, and Non-negative constraint (version \ref{v5}, figure \ref{224-avg-depthwise-const})
  \item Base model with Averaging, Depthwise Convolutional layer, Non-negative constraint, and 50\% Dropout (version \ref{v6}, figure \ref{224-avg-depthwise-const-dp})
  \item Base model with GWAP (version \ref{v7}, \cite{qiu2018global})
\end{itemize}

\begin{figure}[!ht]
\centering
\subfloat[DenseNet121 validation accuracy per epochs]{\label{224-imagenet-fig-densenet121}\includegraphics[width=0.5\linewidth]{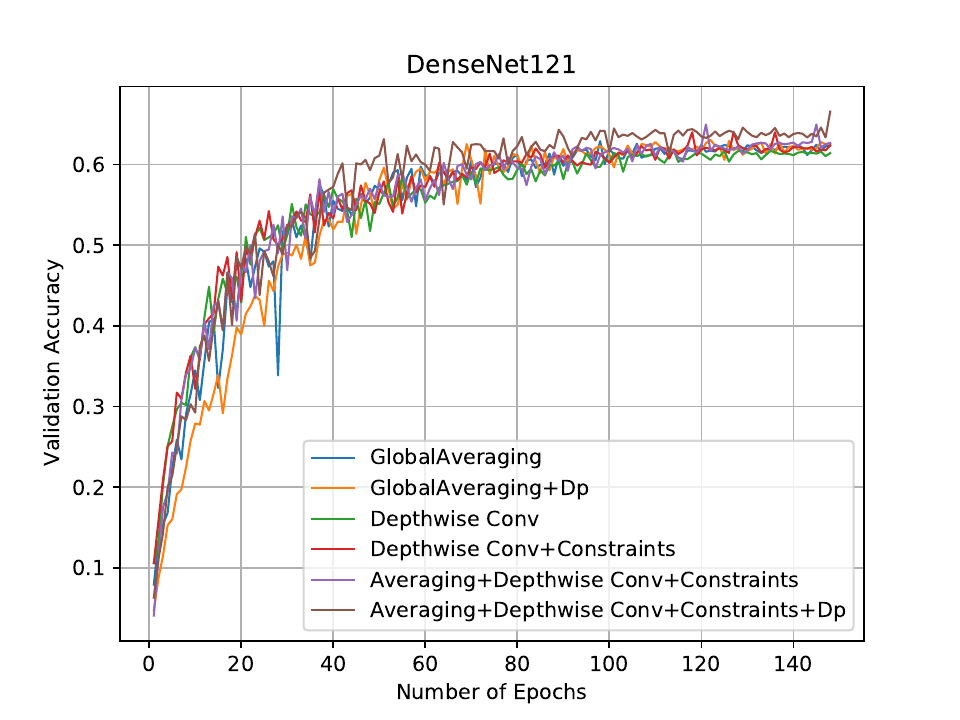}}
\subfloat[ResNet50 validation accuracy per epochs]{\label{224-imagenet-fig-resnet50}\includegraphics[width=0.5\linewidth]{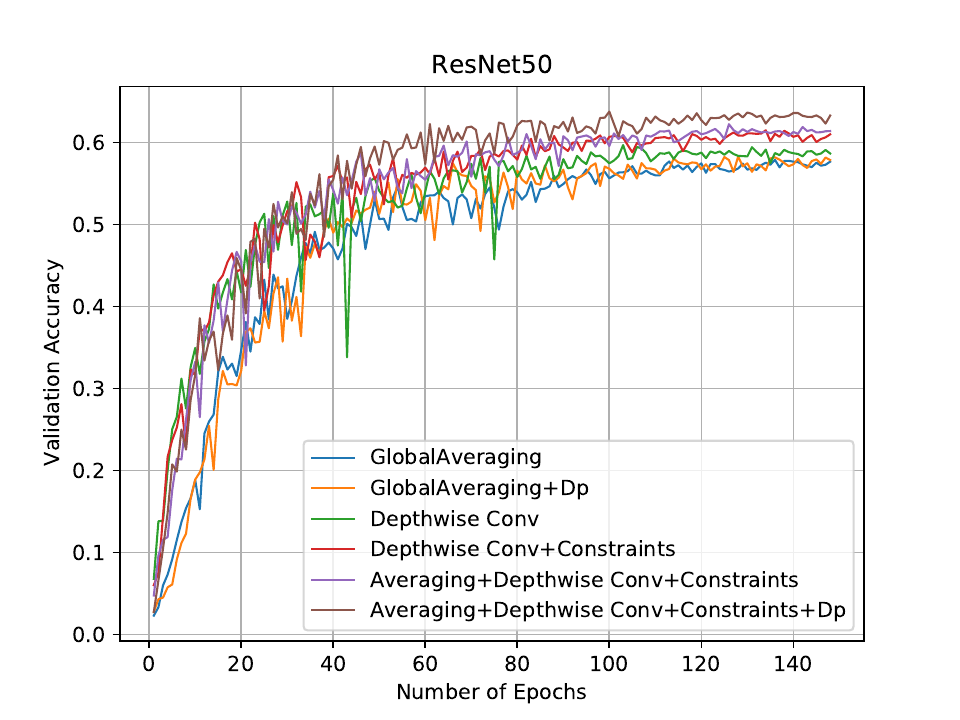}}
\newline
\subfloat[Xception validation accuracy per epochs]{\label{224-imagenet-fig-xception}\includegraphics[width=1\linewidth]{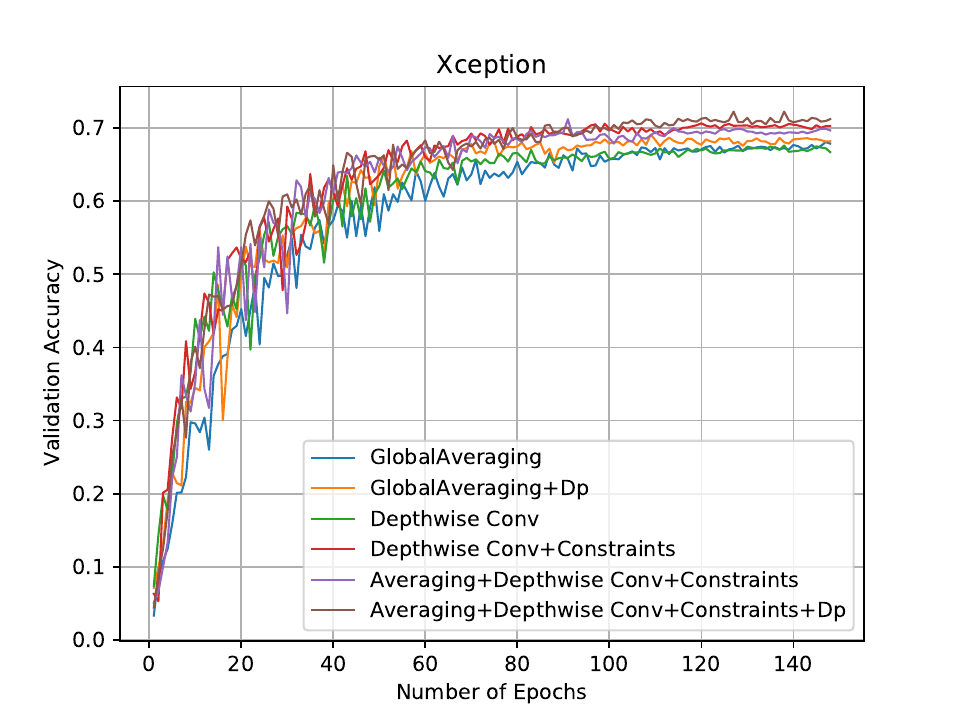}}
\newline
\caption{This figure shows the validation accuracy of the trained models on the sub-ImageNet dataset. For each family, six versions of models (based on classification techniques) have been trained and evaluated to clarify our proposed architecture's performance. The images were resized to 224×224 pixels. The Averaging layer's kernel size was 2×2, and DP refers to a dropout layer with 50\% neurons reduction.}
\label{224-imagenet-fig}
\end{figure}

The architectures of the implemented models are depicted in the referenced figures. The results of this stage of our work are expressed in Table. \ref{224-imagenet} and figure \ref{224-imagenet-fig}

\begin{table*}[!ht]
\centering
\large
\caption{In this table, Top-1 and Top-5 accuracies and the number of parameters (weights) are reported for the trained models on the sub-ImageNet dataset. For each model, seven different versions based on the classification layers have been investigated. Model+GAP is the default classification model that uses the Global Average pooling layer (figure \ref{224-gap}). Model+GAP+DP is the same as Model+GAP but utilizes a 50\% dropout layer (figure \ref{224-gap-dp}).
Model+Depthwise Conv stands for using our depthwise Convolution layer for spatial data analysis (figure \ref{224-depthwise}). Model +Depthwise Conv+Constraints uses our depthwise convolutional layer alongside non-negative constraint (figure \ref{224-depthwise-const}). Model+Avg+Depthwise Conv+Constraints shows the usage of a 2×2 average pooling before the depthwise convolution layer with constraints (figure \ref{224-avg-depthwise-const}). Model+Avg+Depthwise Conv+Constraints+DP applies a 50\% dropout layer to Model+Avg+Depthwise Conv+Constraints (figure \ref{224-avg-depthwise-const-dp}), and Model+GWAP is the implemented version of this reserach \cite{qiu2018global}}
\begin{adjustbox}{width=1\linewidth}
\begin{tabular}{|l|l|l|l|}
\hline
Model name  & Model + Classification type                   & Top-1 accuracy  & Top-5 accuracy  \\ \hline
            & Xception+GAP                                  & 0.6806          & 0.8769          \\ \cline{2-4} 
            & Xception+GAP+DP                               & 0.6909          & 0.8920          \\ \cline{2-4} 
            & Xception+Depthwise Conv                       & 0.6760          & 0.8751          \\ \cline{2-4} 
Xception    & Xception+Depthwise Conv+Constraints           & 0.7063          & 0.8929          \\ \cline{2-4} 
            & Xception+Avg+Depthwise Conv+Constraints       & 0.7120          & 0.8989          \\ \cline{2-4} 
            & Xception+Avg+Depthwise Conv+Constraints+DP    & \textbf{0.7223} & \textbf{0.9106} \\ \cline{2-4} 
            & Xception+GWAP                                 & 0.6307          & 0.8491          \\ \hline
            & ResNet50+GAP                                  & 0.5789          & 0.8066          \\ \cline{2-4} 
            & ResNet50+GAP+DP                               & 0.5829          & 0.8194          \\ \cline{2-4} 
            & ResNet50+Depthwise Conv                       & 0.5969          & 0.8240          \\ \cline{2-4} 
ResNet50    & ResNet50+Depthwise Conv+Constraints           & 0.6149          & 0.8506          \\ \cline{2-4} 
            & ResNet50+Avg+Depthwise Conv+Constraints       & 0.6229          & 0.8634          \\ \cline{2-4} 
            & ResNet50+Avg+Depthwise Conv+Constraints+DP    & \textbf{0.6377} & \textbf{0.8677} \\ \cline{2-4} 
            & ResNet50+GWAP                                 & 0.5365          & 0.7811          \\ \hline
            & DenseNet121+GAP                               & 0.6289          & 0.8508          \\ \cline{2-4} 
            & DenseNet121+GAP+DP                            & 0.6303          & 0.8520          \\ \cline{2-4} 
            & DenseNet121+Depthwise Conv                    & 0.6171          & 0.8446          \\ \cline{2-4} 
DenseNet121 & DenseNet121+Depthwise Conv+Constraints        & 0.6399          & 0.8651          \\ \cline{2-4} 
            & DenseNet121+Avg+Depthwise Conv+Constraints    & 0.6495          & 0.8694          \\ \cline{2-4} 
            & DenseNet121+Avg+Depthwise Conv+Constraints+DP & \textbf{0.6654} & \textbf{0.8740} \\ \cline{2-4} 
            & DenseNet121+GWAP                              & 0.5828          & 0.8239          \\ \hline
\end{tabular}
\end{adjustbox}

\label{224-imagenet}
\end{table*}

Based on the obtained results of figure \ref{224-imagenet-fig}, and Table. \ref{224-imagenet}, it is visible that using a single Depthwise convolutional layer shows higher validation accuracy in the first epochs of training but falls into overfitting wholes in the following epochs. Applying a non-negative constrain to the depthwise convolutional layer will decrease the overfitting effects, and the model's classification accuracy rises. Therefore, the models' results with a depthwise convolutional layer and constraints have been significantly improved than using the GAP layer.

 Utilizing an averaging layer before the depthwise convolutional layer will still reduce overfitting and enhance learning quality. We also investigated the usage of Dropout on our final proposed model to show that our architecture can also be improved by using dropouts and is not dependent on all features.

Our final proposed architecture which is called Wise-SrNet (figure \ref{224-avg-depthwise-const-dp}) increased the Top-1 accuracy of Xception, ResNet50, and DenseNet121 models with GAP and dropout layers by 3.14\%, 5.48\%, and 3.51\% respectively. We have also reported the number of weights of each model in Table. \ref{224-imagenet} to clarify that our architectures do not increase computational costs compared to the GAP layer.

Here, this question may raise why the reported Top-1 classification accuracy of models with the GAP layer is lower than what their developers reported on the whole ImagNet dataset, and the Top-5 accuracy of our models is higher? There are two answers to this question:

\begin{enumerate}
\item ImageNet dataset includes around 1000-1500 training images and 50 validation images for each class, but our sub-ImageNet dataset consists of 500 training images and 50 validation images for each class. As our training images in almost one-third of the training images of the ImageNet dataset and the number of validation images is the same, it is clear that we can not reach the results that the developers of the same models with the GAP layer reached on the whole ImageNet dataset.
\item The developers of deep convolutional models like Xception and ResNet had access to powerful hardware and lots of GPUs, allowing them to train the models on ImageNet with high batch size (usually the batch size is set to 256). But as we did not have many GPUs available, we could not train our models on the whole ImageNet dataset. Due to the unavailability of more GPU ram space, we were forced to set the batch size to 70 for 224×224 resolution images. Having fewer batch sizes will decrease the learning quality and efficiency for sure.
\end{enumerate}

Because of these two reasons, our obtained Top-1 classification accuracy of using the GAP layer is lower than what is reported by the default models' authors. But as we had fewer classes than the whole ImageNet dataset (70 vs. 1000), the Top-5 accuracy of our trained models (with GAP layer) is higher than the default models with the GAP layer.

Our wish is to show that using our architecture will significantly increase the classification accuracy, and our results prove it. Other researchers that have access to more GPUs can train the deep convolutional models using our architecture on the whole ImageNet dataset and report the new and improved results. We think our methods can create a great improvement in the classification benchmarks.

\subsubsection{Intel Image Classification Challenge}
\label{312}

Intel Image Classification Dataset \cite{intel} consists of 14034 training images and 3000 test images belonging to 6 classes. The default resolution of this dataset's images is 150×150 pixels, which we resized to 224×224. For this dataset, we chose Xception \cite{chollet2017xception}, DenseNet169 \cite{huang2018densely}, and ResNet50 \cite{he2015deep} models for further experiments.
We trained the models for 220 epochs. 

On this dataset, we evaluated three versions of each model:

\begin{itemize}
  \item Base model with GAP layer (version \ref{v1}, figure \ref{224-gap})
  \item Base model with Depthwise Convolutional layer and Non-negative constraint (version \ref{v4}, figure \ref{224-depthwise-const})
  \item Base model with Averaging, Depthwise Convolutional layer and Non-negative constraint (version \ref{v5}, figure \ref{224-avg-depthwise-const})
\end{itemize}

\begin{figure}[!ht]
\centering
\subfloat[DenseNet169 validation accuracy per epochs]{\label{224-intel-val-densenet169}\includegraphics[width=0.5\linewidth]{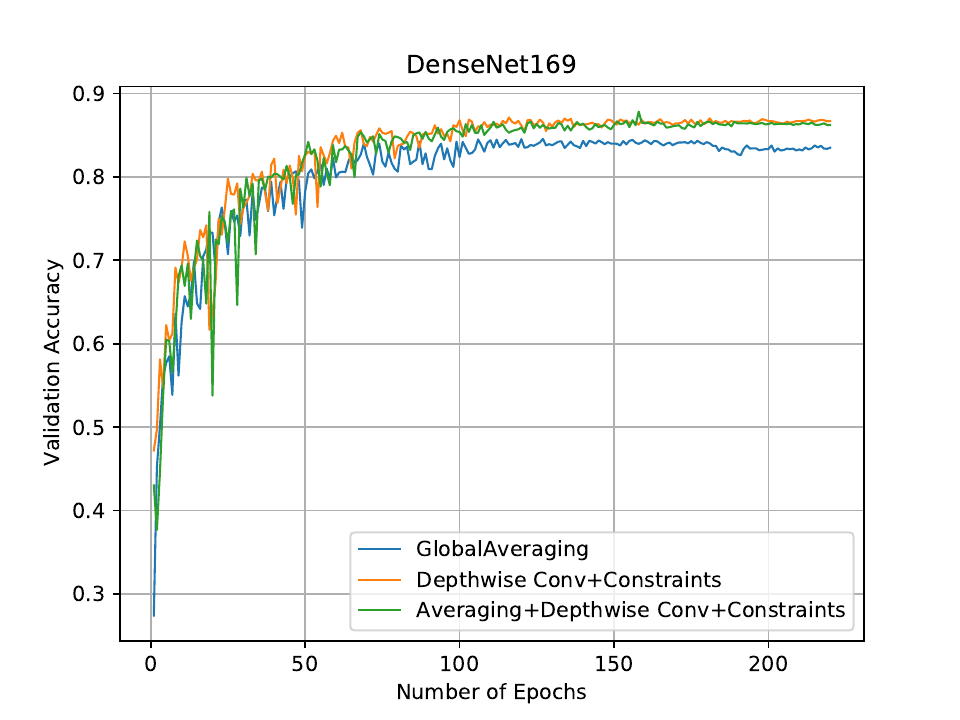}}
\subfloat[DenseNet169 training accuracy per epochs]{\label{224-intel-acc-densenet169}\includegraphics[width=0.5\linewidth]{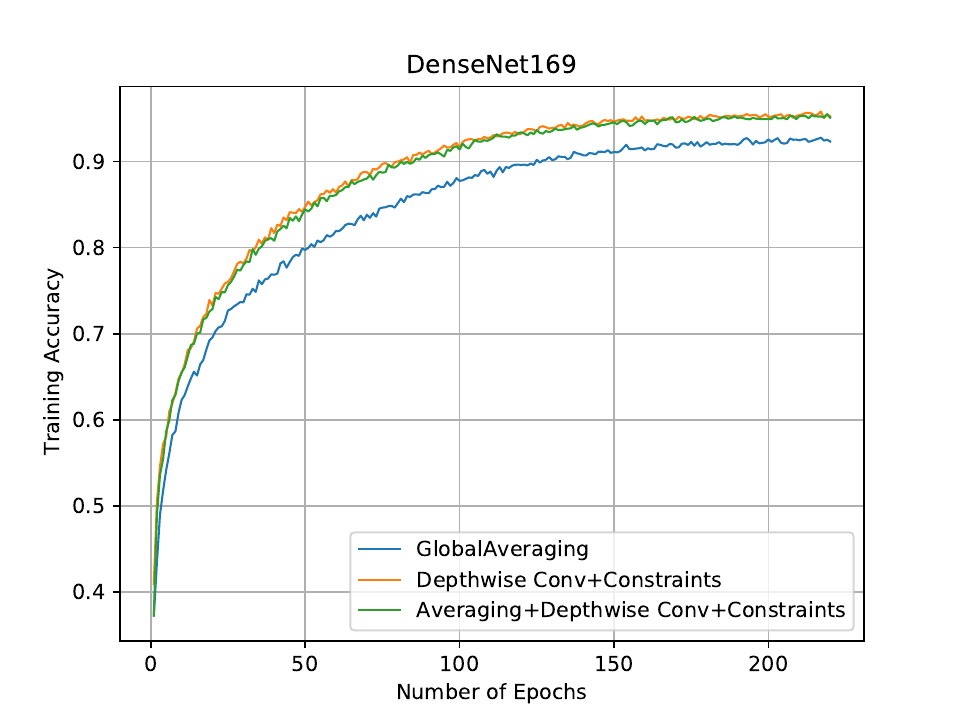}}
\newline
\subfloat[ResNet50 validation accuracy per epochs]{\label{224-intel-val-resnet50}\includegraphics[width=0.5\linewidth]{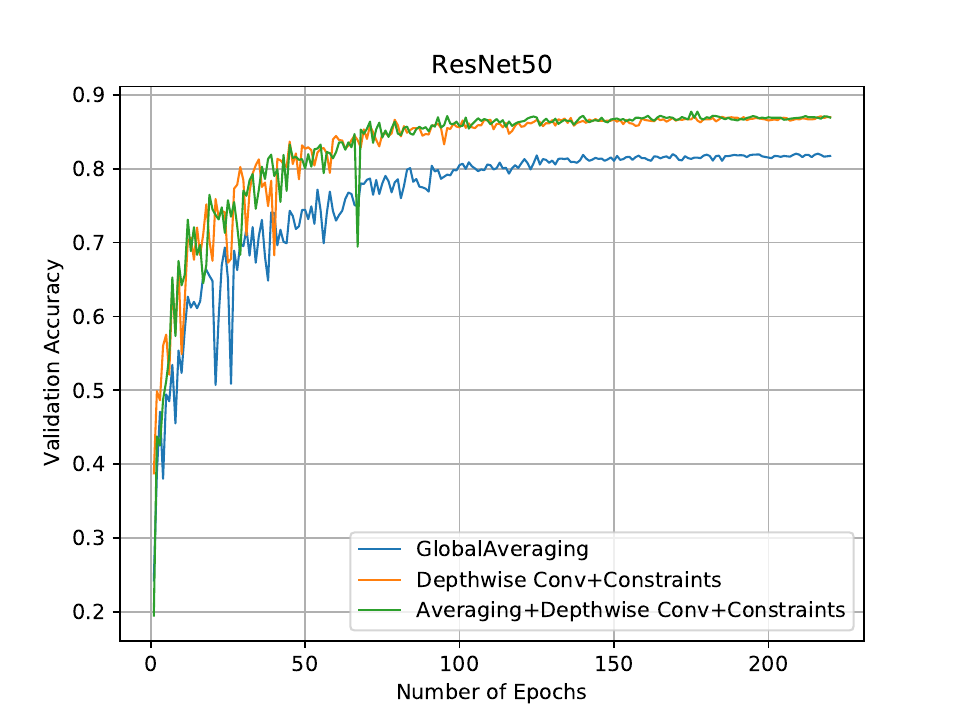}}
\subfloat[ResNet50 training accuracy per epochs]{\label{224-intel-acc-resnet50}\includegraphics[width=0.5\linewidth]{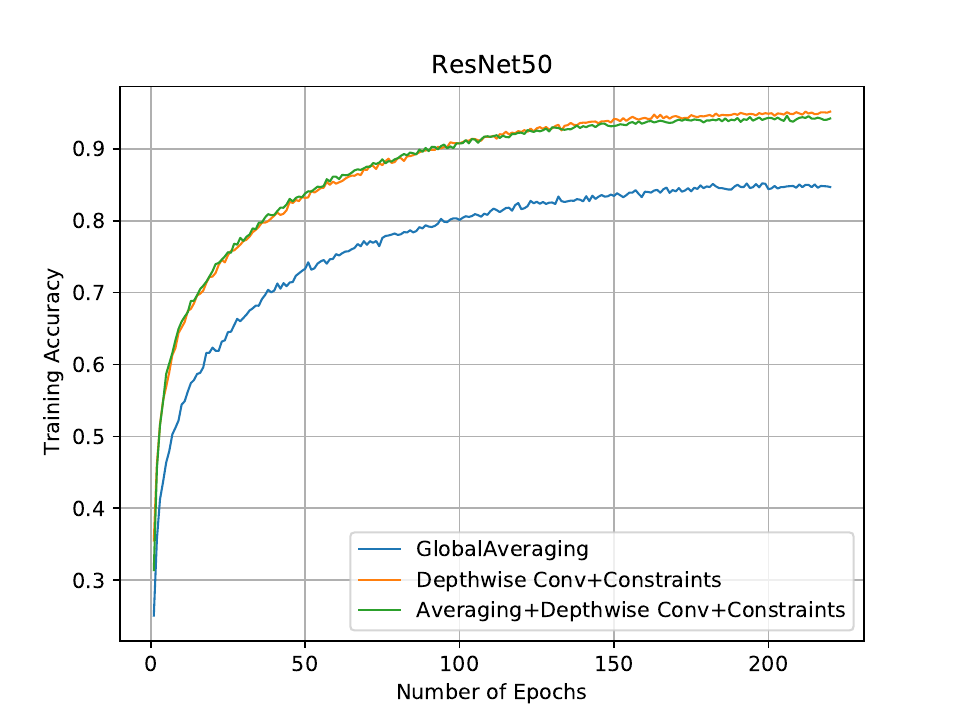}}
\newline
\subfloat[Xception validation accuracy per epochs]{\label{224-intel-val-xception}\includegraphics[width=0.5\linewidth]{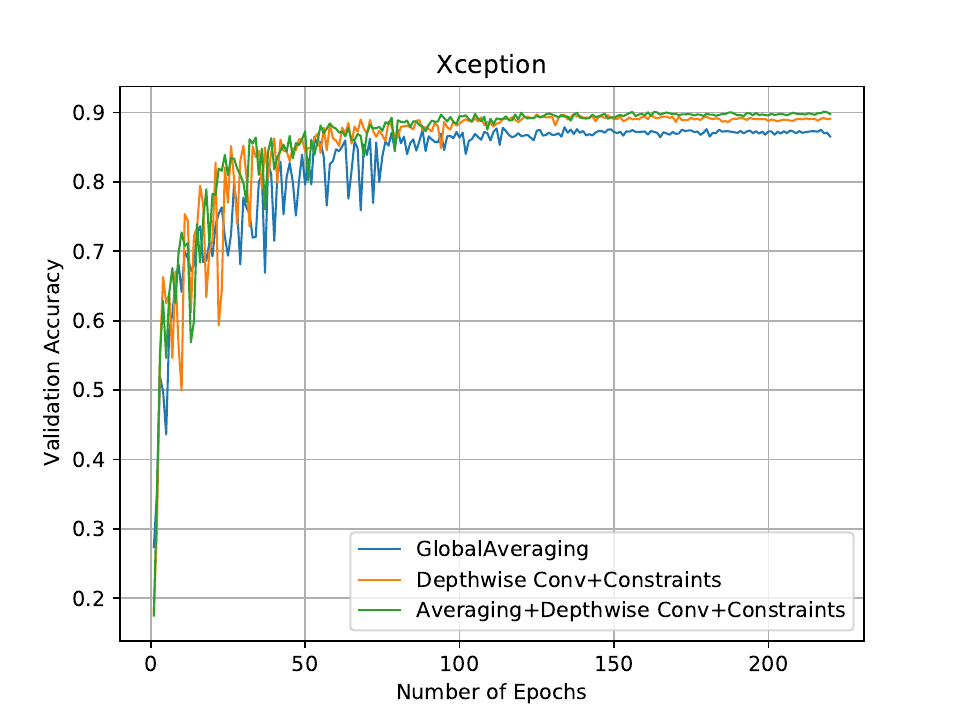}}
\subfloat[Xception training accuracy per epochs]{\label{224-intel-acc-xception}\includegraphics[width=0.5\linewidth]{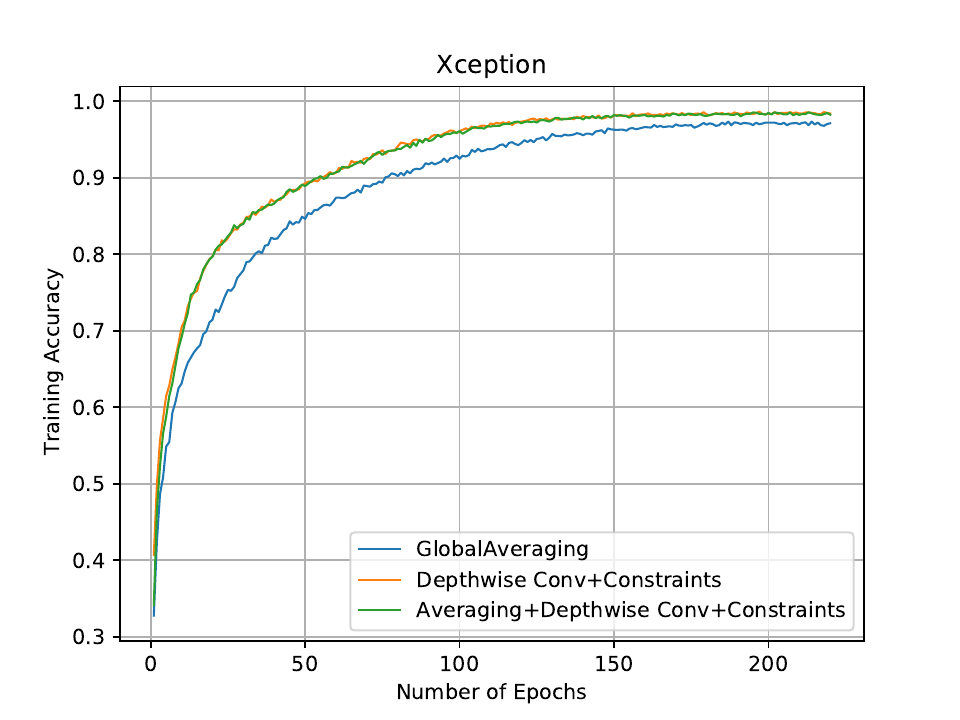}}

\caption{In this figure, the validation and training accuracies of the trained models on the Intel Image Classification dataset for 220 epochs are presented. The images were resized to 224×224 pixels. The Averaging layer's kernel size was 2×2.}
\label{224-intel-all}
\end{figure}

The results of our trained models on the Intel Image Classification dataset are detailed and depicted in Table. \ref{224-intel}, and figure \ref{224-intel-all}.

\begin{table*}[!hb]
\centering
\large
\caption{This table shows the results of the trained models on the Intel Image Classification dataset. Xception, ResNet50, and DenseNet169 models were selected for evaluating our architecture. For each of these modes, three versions were validated. Model+GAP is the default classification model that uses the Global Average pooling layer (figure \ref{224-gap}). Model +Depthwise Conv+Constraints uses our depthwise convolutional layer alongside non-negative constraint (figure \ref{224-depthwise-const}). Model+Avg+Depthwise Conv+Constraints shows the usage of an average pooling before the depthwise convolution layer with constraints (figure \ref{224-avg-depthwise-const}).}
\begin{tabular}{|l|l|l|l|}
\hline
Model name  & Model + Classification type                & Parameters & Top-1 accuracy  \\ \hline
            & Xception+GAP                               & 20,873,774 & 0.8787          \\ \cline{2-4} 
Xception    & Xception+Depthwise Conv+Constraints        & 20,976,174 & 0.8977          \\ \cline{2-4} 
            & Xception+Avg+Depthwise Conv+Constraints    & 20,894,254 & \textbf{0.9013} \\ \hline
            & ResNet50+GAP                               & 23,600,006 & 0.8203          \\ \cline{2-4} 
ResNet50    & ResNet50+Depthwise Conv+Constraints        & 23,702,406 & 0.8713          \\ \cline{2-4} 
            & ResNet50+Avg+Depthwise Conv+Constraints    & 23,620,486 & \textbf{0.8773} \\ \hline
            & DenseNet169+GAP                            & 12,652,870 & 0.8457          \\ \cline{2-4} 
DenseNet169 & DenseNet169+Depthwise Conv+Constraints     & 12,736,070 & 0.8713          \\ \cline{2-4} 
            & DenseNet169+Avg+Depthwise Conv+Constraints & 12,669,510 & \textbf{0.8783} \\ \hline
\end{tabular}

\label{224-intel}
\end{table*}





As mentioned, three different classification techniques were investigated for each of these networks: The GAP layer, the Depthwise convolutional layer with constraints, and Averaging with the depthwise convolutional layer and the non-negative constraint.

Figures \ref{224-intel-acc-densenet169}, \ref{224-intel-acc-resnet50}, and \ref{224-intel-acc-xception} present the training accuracy of the trained models on this dataset. The figures clearly show that the usage of a depthwise convolutional layer with non-negative constraints will enhance the training process. This figure's important point is revealing that the ResNet50 network will show much better performance when using our architecture than the GAP layer.

Figures \ref{224-intel-val-densenet169}, \ref{224-intel-val-resnet50}, and \ref{224-intel-val-xception} and Table. \ref{224-intel} show the Top-1 accuracy of the trained models. The information on this table and figures also prove the fact that our architecture can significantly improve classification results. Using averaging before the depthwise convolution layer can also affect the classification process by preventing models from overfitting.

Our proposed architecture increases the Top-1 accuracy of Xception, ResNet50, and DenseNet169 models by 2.26\%, 5.7\%, and  3.26\% on Intel Image Classification dataset, respectively. Reported parameters show that the models with our architecture contain almost the same number of parameters as those models with the GAP layer.

 It is noteworthy that dropout layers can also improve the classification results, but our goal in this research is to show that our architecture performs better than old GAP methods. However, we investigated the use of dropout on other datasets.

\subsubsection{MIT Indoors Scenes}
\label{313}

MIT Indoors Scenes dataset \cite{quattoni2009recognizing} includes 67 classes of images of different scenes and views. This dataset is made of 5360 training images and 1340 validation images. The default resolution of its pictures is mixed, but usually, the images have good resolution. At this stage, we resized all the images to 224×224 for our experiments.
Xception, DenseNet169, and ResNet50 networks were chosen as our reference models in this phase. For each of these networks, five models with different classification layers were evaluated:

\begin{itemize}
  \item Base model with GAP layer (version \ref{v1}, figure \ref{224-gap})
  \item Base model with GAP layer and 50\% Dropout (version \ref{v2}, figure \ref{224-gap-dp})
  \item Base model with Depthwise Convolutional layer and, and Non-negative constraint (version \ref{v4}, figure \ref{224-depthwise-const})
  \item Base model with Averaging, Depthwise Convolutional layer, and Non-negative constraint (version \ref{v5}, figure \ref{224-avg-depthwise-const})
  \item Base model with Averaging, Depthwise Convolutional layer, Non-negative constraint and 50\% Dropout (version \ref{v6}, figure \ref{224-avg-depthwise-const-dp})
\end{itemize}

\begin{figure}[!ht]
\centering
\subfloat[DenseNet169 validation accuracy per epochs]{\label{224-mit-fig-densenet169}\includegraphics[width=0.5\linewidth]{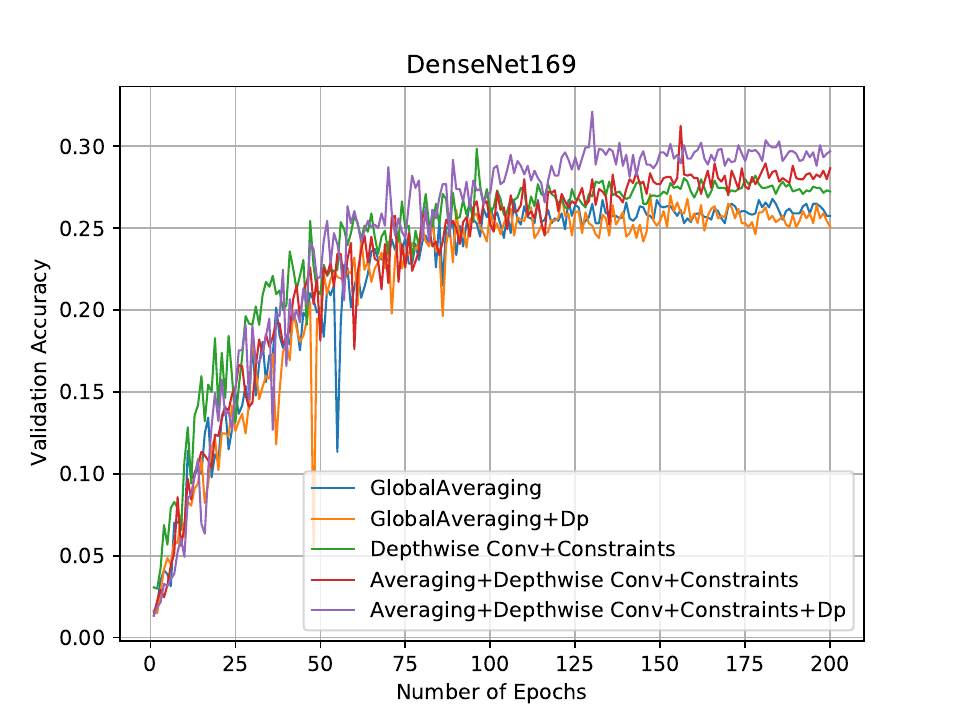}}
\subfloat[ResNet50 validation accuracy per epochs]{\label{224-mit-fig-resnet50}\includegraphics[width=0.5\linewidth]{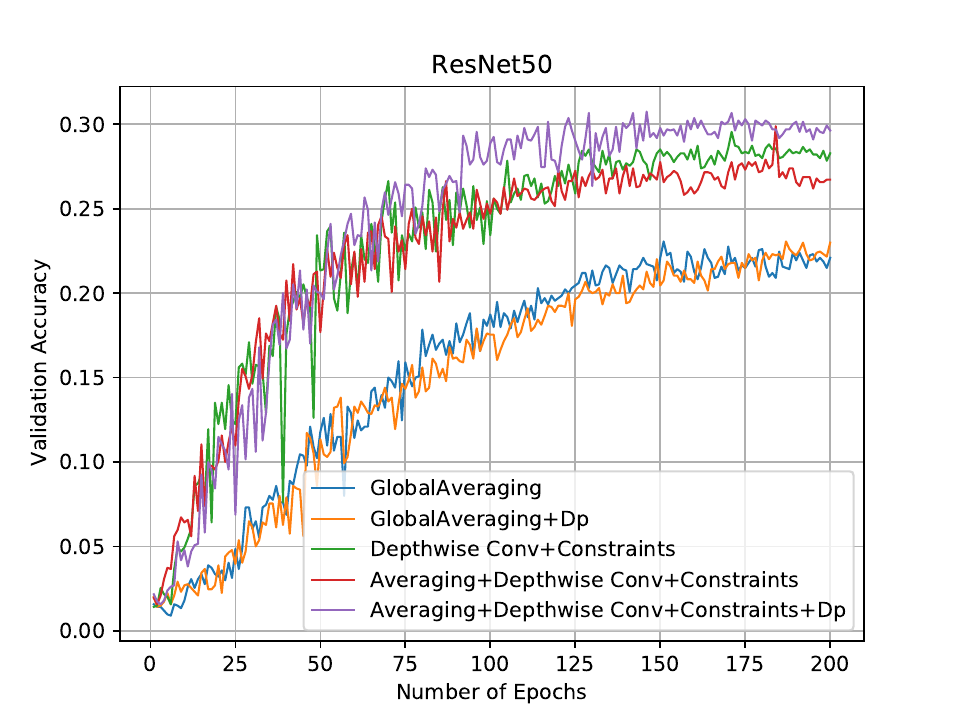}}
\newline
\subfloat[Xception validation accuracy per epochs]{\label{224-mit-fig-xception}\includegraphics[width=1\linewidth]{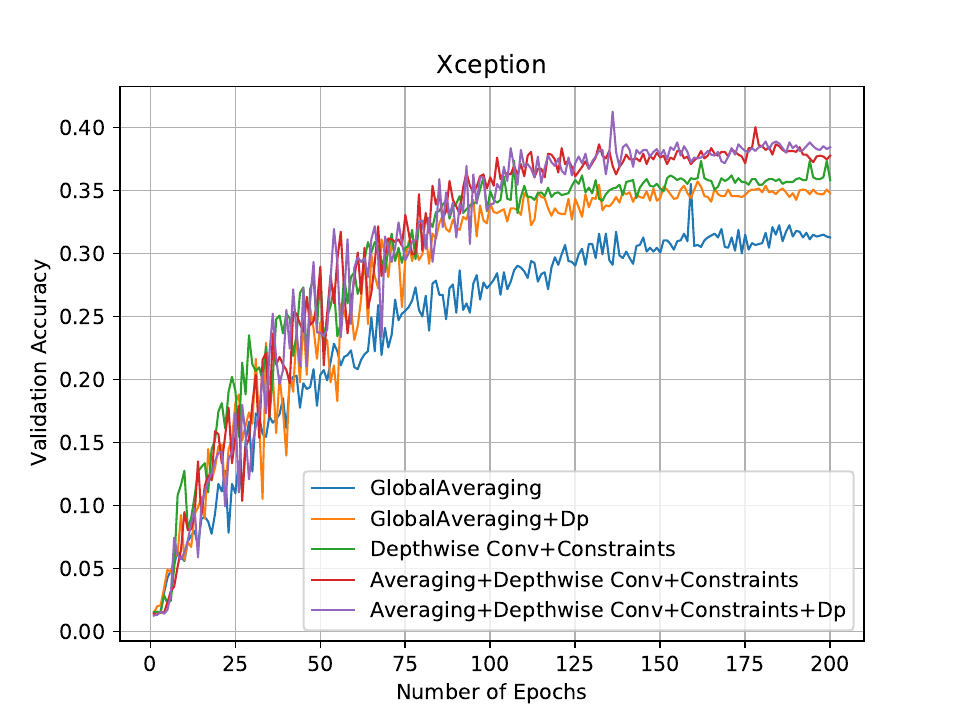}}

\caption{In this figure, the validation accuracy of the trained models with different classification layers on the MIT Indoors Scenes dataset for 200 epochs is presented. The images were resized to 224×224 pixels. The Averaging layer's kernel size was 2×2, and DP refers to a dropout layer with 50\% neurons reduction.}
\label{224-mit-fig}
\end{figure}

Table. \ref{224-mit} and figure \ref{224-mit-fig} show our obtaned results on this dataset.

\begin{table*}
\centering
\large
\caption{In this table, the Top-1 accuracy and Top-5 accuracy of trained models with different classification layers are presented. The number of each model's parameters is also mentioned for revealing the models' computational costs.}
\begin{adjustbox}{width=1\linewidth}
\begin{tabular}{|l|l|l|l|l|}
\hline
Model name  & Model + Classification type                   & Parameters & Top-1 accuracy  & Top-5 accuracy  \\ \hline
            & Xception+GAP                                  & 20,998,763 & 0.3552          & 0.6485          \\ \cline{2-5} 
            & Xception+GAP+DP                               & 20,998,763 & 0.3571          & 0.6515          \\ \cline{2-5} 
Xception    & Xception+Depthwise Conv+Constraints           & 21,101,163 & 0.3734          & 0.6622          \\ \cline{2-5} 
            & Xception+Avg+Depthwise Conv+Constraints       & 21,019,243 & 0.4003          & 0.6943          \\ \cline{2-5} 
            & Xception+Avg+Depthwise Conv+Constraints+DP    & 21,019,243 & \textbf{0.4125} & \textbf{0.7011} \\ \hline
            & ResNet50+GAP                                  & 23,724,995 & 0.2306          & 0.5179          \\ \cline{2-5} 
            & ResNet50+GAP+DP                               & 23,724,995 & 0.2313          & 0.5321          \\ \cline{2-5} 
ResNet50    & ResNet50+Depthwise Conv+Constraints           & 23,827,395 & 0.2955          & 0.5582          \\ \cline{2-5} 
            & ResNet50+Avg+Depthwise Conv+Constraints       & 23,745,475 & 0.2988          & 0.5799          \\ \cline{2-5} 
            & ResNet50+Avg+Depthwise Conv+Constraints+DP    & 23,745,475 & \textbf{0.3075} & \textbf{0.5806} \\ \hline
            & DenseNet169+GAP                               & 12,754,435 & 0.2679          & 0.5328          \\ \cline{2-5} 
            & DenseNet169+GAP+DP                            & 12,754,435 & 0.2709          & 0.5478          \\ \cline{2-5} 
DenseNet169 & DenseNet169+Depthwise Conv+Constraints        & 12,837,635 & 0.2986          & 0.5588          \\ \cline{2-5} 
            & DenseNet169+Avg+Depthwise Conv+Constraints    & 12,771,075 & 0.3125          & 0.5590          \\ \cline{2-5} 
            & DenseNet169+Avg+Depthwise Conv+Constraints+DP & 12,771,075 & \textbf{0.3212} & \textbf{0.5981} \\ \hline
\end{tabular}
\end{adjustbox}

\label{224-mit}
\end{table*}

The validation data of Table .\ref{224-mit} and figure \ref{224-mit-fig} illuminates this idea that applying depthwise convolutional layer with constraints will result in achieving higher classification accuracy. Our final architecture \ref{224-avg-depthwise-const-dp} increases the Top-1 validation accuracy of Xception, ResNet50, and DenseNet169 models with GAP layer by 5.54\%, 7.62\%, 5.03\% respectively.

Now we wish to explain why the classification results on the MIT Indoors Scenes are not high? This dataset contains 5360 training images of 67 classes, which means there are only 80 images for each class. This number is too low for a model to learn from scratch. As the number of total classes is considerable, so the models can not achieve better results, and the learning can not be executed appropriately. Another point is that we did not use any transfer learning, so the models are being trained from scratch, and based on all these reasons, the models can not converge to better learning points.

 Of course, there can be many modifications to improve the results, like increasing the resolution of images, using transfer learning, which we will implement in Section. \ref{32}, and we will reach promising results.

\subsection{Experiments on Images with 512×512 Pixels Resolution}
\label{32}

In this stage of our work, we wish to show interesting findings. In computer vision tasks, utilizing larger input images usually means achieving higher accuracy. Images' resolution will shrink when passing through the deep convolutional model layers, and at the end, a compressed feature map will be delivered. In this while, some image information, especially information related to smaller objects and areas, may become lost or removed.
 So, when the input image is larger, the objects and areas inside the image will be wider; therefore, the possibility of removing important information will be lower, and the models show higher precision.

When training on the Imagenet dataset, it is common to resize the images to 224×224 pixels because this is a large dataset, and working with larger images needs much more hardware and will also be time-consuming and difficult. But, in other cases, many developers chose to work with larger images to obtain better results. Besides, in many medical image analysis cases, as the infections or hotspots exist in small areas, it forces the researchers to use larger images to achieve acceptable results.

This section shows the effect of our architecture when the input images' size is 512×512. Here, we only work with the MIT Indoors Scenes dataset because the original images of this dataset are in good resolution, and the number of classes of this dataset is suitable for our work.

In this section, Unlike the previous steps, we used transfer learning from the ImageNet pre-trained weights to accelerate training speed and convergence. 
A critical point that we mentioned before is that using transfer learning from pre-trained weights of models trained on the Global Average Pooling (GAP) layer will make our comparison unfair. 
The available weights are produced based on models with the GAP layer, which have been trained on 224×224 images for many epochs and converged to the optimum point.
Training the models with the GAP layer will affect all the models' weights; somehow, the model will learn to export the final feature map suitable for global averaging. Suppose we apply these pre-trained weights at the beginning of our training for comparing our architecture with the GAP layer.  As the pre-trained weights are produced for using the GAP layer on 224×224 images, then the comparison will not be correct. Because in that way, the models will know how to deal with the same size of images completely suitable for global averaging, but this condition does not exist for our architecture.

But in current condition that the images are in 512×512 resolution and the pre-trained weights are produced for 224×224 images, they will not be very similar. Although these weights are created for models with the GAP layer, we will prove that our architecture still works much better than the GAP layer, even with transfer learning from the explained weights.

We used the same data augmentation techniques as reported in Table. \ref{data augment}. All the implemented parameters are also expressed in Table. \ref{512-param}. We changed the batch size value from 70 to 15 due to the need for more RAM space caused by using larger images. We continued the training process for 130 epochs, which is less than 220 epochs used for training on the same dataset for 224×224 images as we applied transfer learning at the beginning of the training, which makes the models converge in fewer epochs.
Other parameters are almost the same.

MIT Indoors Scenes dataset has been fully described in section \ref{313}. In this section we selected Xception \cite{chollet2017xception}, ResNet50V2 \cite{he2016identity}, and DenseNet201 \cite{huang2018densely} for training and investigation. In this phase, we do not wish to compare our different proposed architectures because the results of previous steps prove that our final proposed architectures (Wise-SrNet, figures \ref{224-avg-depthwise-const-dp}, and \ref{224-avg-depthwise-const}) are more robust than others.

Most of the deep convolutional neural networks present a feature map with a kernel size of 1/32 times of input image resolution. When the image size is 512×512, the extracted feature map kernel size will be 16×16. The channel size varies  for different models; Xception and ResNet families produce 2048 channels, and DenseNet121, DenseNet169, and DenseNet201 models generate 1024, 1664, and 1920 channels, respectively.

Now, if we consider working with the Xception model, the feature map size for an input image with 512×512 resolution will be 16×16×2048, while the 224×224 images would output a feature map with 7×7×2048 size. The Global Average Pooling (GAP) layer computes the average of the 16×16=256 values of each channel for 512×512 images, while it has to average between 7×7=49 values of each channel for 224×244 images. Averaging between 256 values can destroy much more spatial information than averaging between 49 spatial values. 

\begin{figure}[!ht]
\centering
\subfloat[Deep convolution model with the GAP layer at its classification section.]{\label{512-gap}\includegraphics[width=\linewidth]{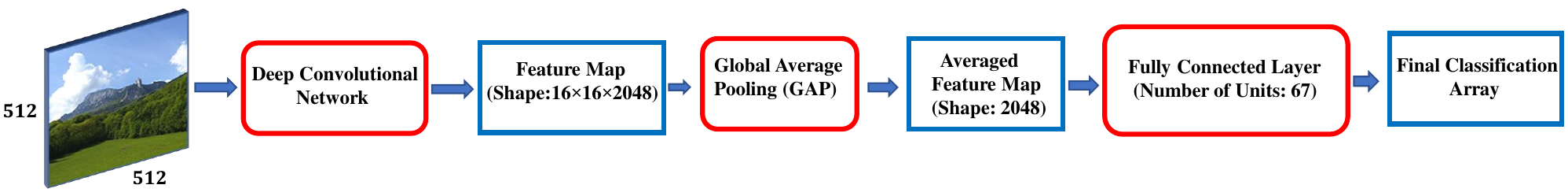}}
\newline
\subfloat[In this architecture, the extracted features are first passed through an average pooling layer for reducing the feature map's size. Then the flattened features will be fed to an FC layer for performing the classification task.]{\label{512-avg-dense}\includegraphics[width=\linewidth]{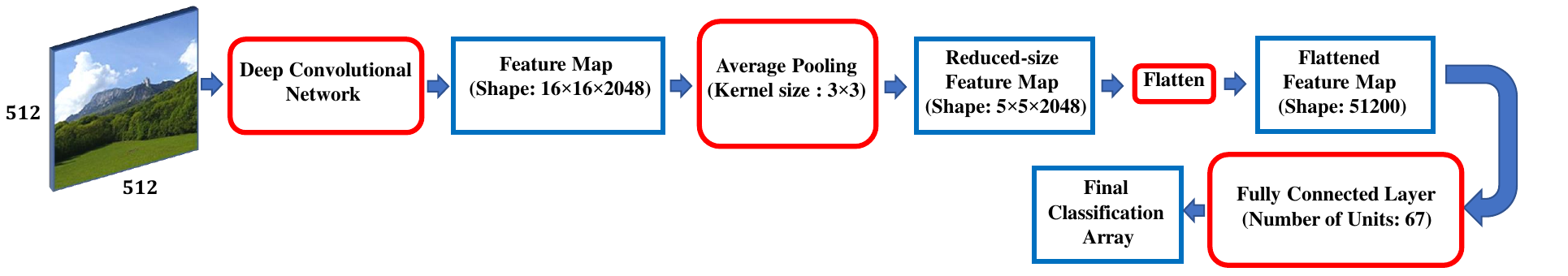}}
\newline

\subfloat[Architecture of a model using depthwise convolutional layer for extracting spatial and channel data with pre-averaging and non-negative constraint to prevent overfitting.]{\label{512-avg-depthwise-const}\includegraphics[width=\linewidth]{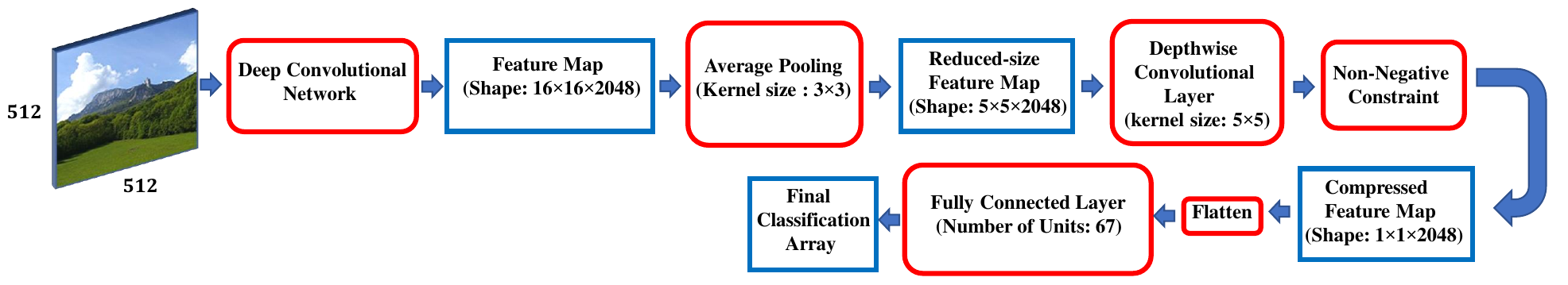}}

\caption{Our designed architectures for working with 512×512 images have been illustrated in these figures. The red boxes are the model computational layers, and the blue boxes show the output arrays.}
\label{512-all}
\end{figure}

This statement shows us that using large images with the GAP layer, especially when there are lots of classes, will not be efficient, and the results can not be proper.
In this phase, we trained, evaluated and compared three version of models for each of the Xception, ResNet50V2, and DenseNet201 networks:

\begin{itemize}
  \item Base model with GAP layer (figure \ref{512-gap})
  \item Base model with Averaging and FC layer (figure \ref{512-avg-dense})
  \item Base model with Averaging, Depthwise Convolutional layer, and Non-negative constraint (figure \ref{512-avg-depthwise-const})
\end{itemize}

The architectures of the base model with GAP layer and Base model with Averaging,  Depthwise Convolutional layer, and Non-negative constraint has been depicted in figure \ref{512-gap}, and figure \ref{512-avg-depthwise-const}, respectively. 

 The input images' resolution is 512×512, so the feature map size in most models (Xception and ResNet) will be 16×16×2048. Accordingly, for the averaging layer, we set the averaging kernel size to 3×3 in such a way that the output feature map size becomes 5×5×2048. Then we set the depthwise convolutional layer kernel size to 5×5; hence, the output will be a 1×1×2048 array that will be converted to a one-dimensional array and fed to the final classification layer as explained in previous sections. There can be many configurations for choosing the kernel size of the average pooling and depthwise convolutional layers and obtaining better results which are not our intention in this research. Here, we wish to prove that using these kinds of architectures instead of the GAP layers will increase classification accuracy.

  Base model with Averaging and FC layer is another architecture we wish to explore in this stage. This architecture is presented in figure \ref{512-avg-dense}. It is inspired from figure \ref{dense}, which shows a classic way for classification, but the difference is that we could not feed the flattened feature map to the final classification fully connected layer. Because when the image size is 512×512, the feature map shape would be 16×16×2048, and the flattened feature map will have 524,288 neurons. If we feed this array to the final FC layer with 67 classes, the number of parameters will increase by 35,127,69 weights! Adding this number of trainable parameters to the model will cause overfitting and learning deficiency. So we attempted to use another approach, as you can see in figure \ref{512-avg-dense}; first, we applied a 3×3 averaging layer to the feature map to produce a 5×5×2048 array, then we flattened it and fed it to the final FC layer. This way, the number of model parameters will not be so high. Our goal of using this architecture is to prove that our proposed architecture can also surpass this classic architecture in both learning efficiency and computational cost. In other words, we want to say that our architecture shows the best performance.

\begin{figure}[!ht]
\centering
\subfloat[DenseNet201 training accuracy per epochs]{\label{512-mit-acc-DenseNet201}\includegraphics[width=0.5\linewidth]{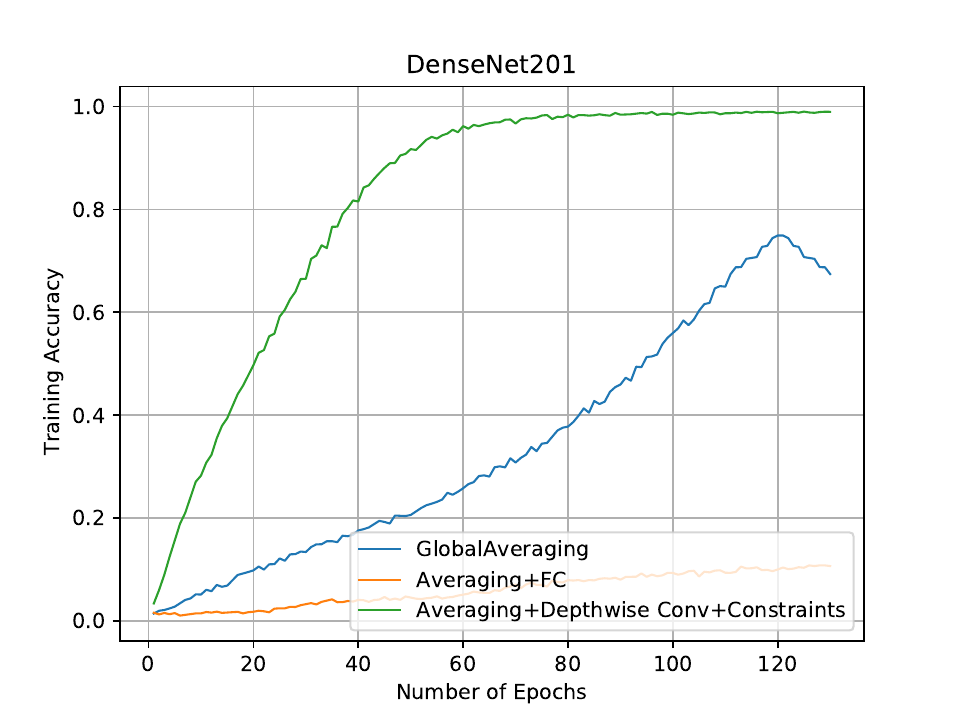}}
\subfloat[DenseNet201 validation accuracy per epochs]{\label{512-mit-val-DenseNet201}\includegraphics[width=0.5\linewidth]{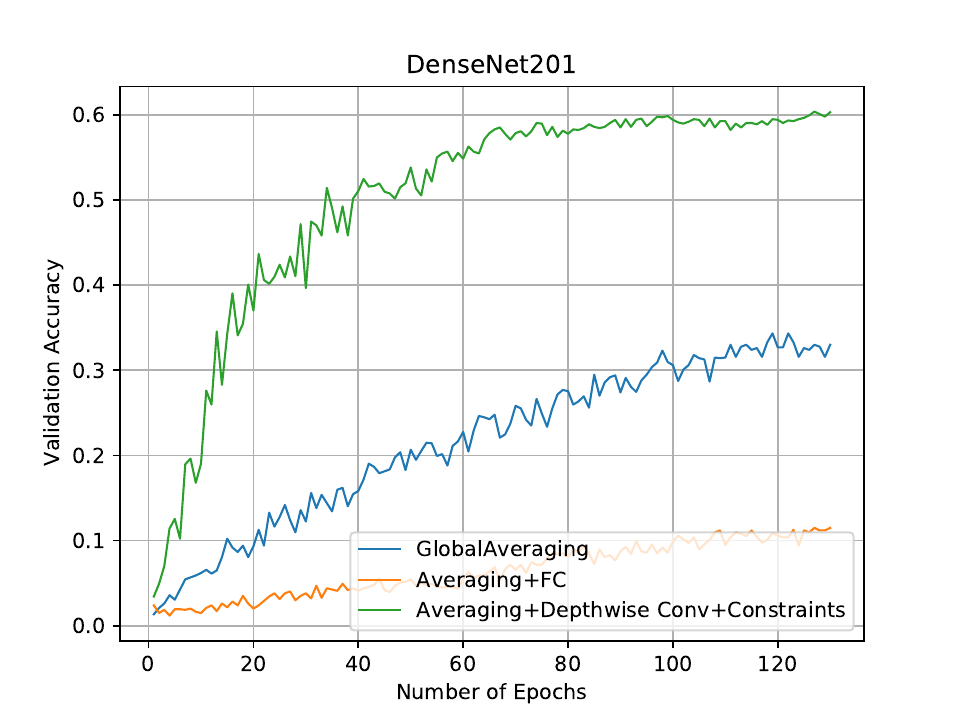}}
\newline
\subfloat[ResNet50V2 training accuracy per epochs]{\label{512-mit-acc-ResNet50V2}\includegraphics[width=0.5\linewidth]{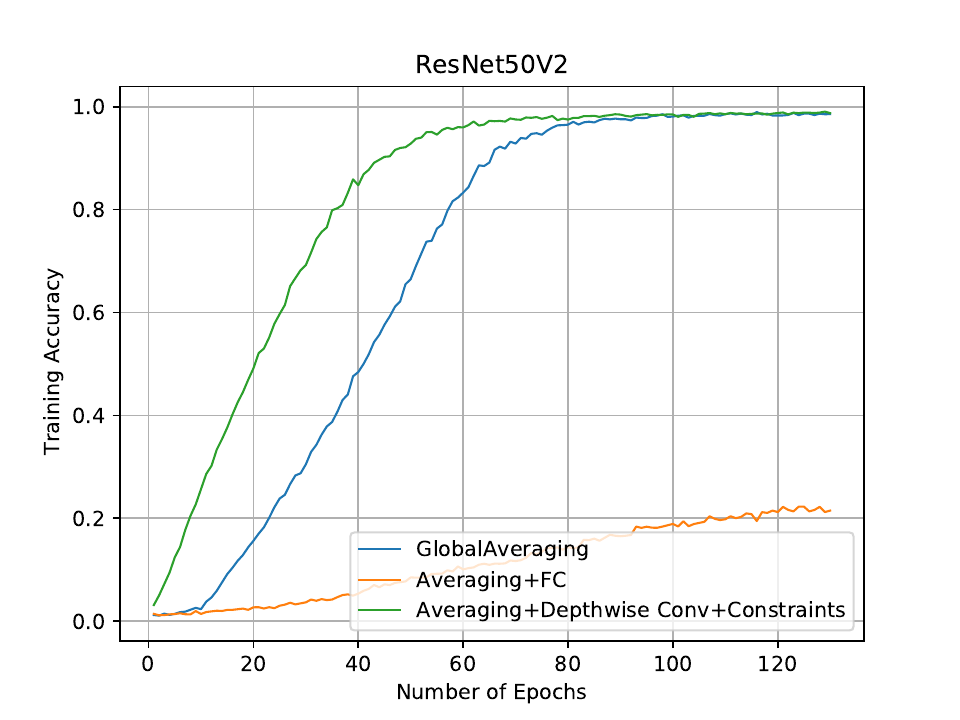}}
\subfloat[ResNet50V2 validation accuracy per epochs]{\label{512-mit-val-ResNet50V2}\includegraphics[width=0.5\linewidth]{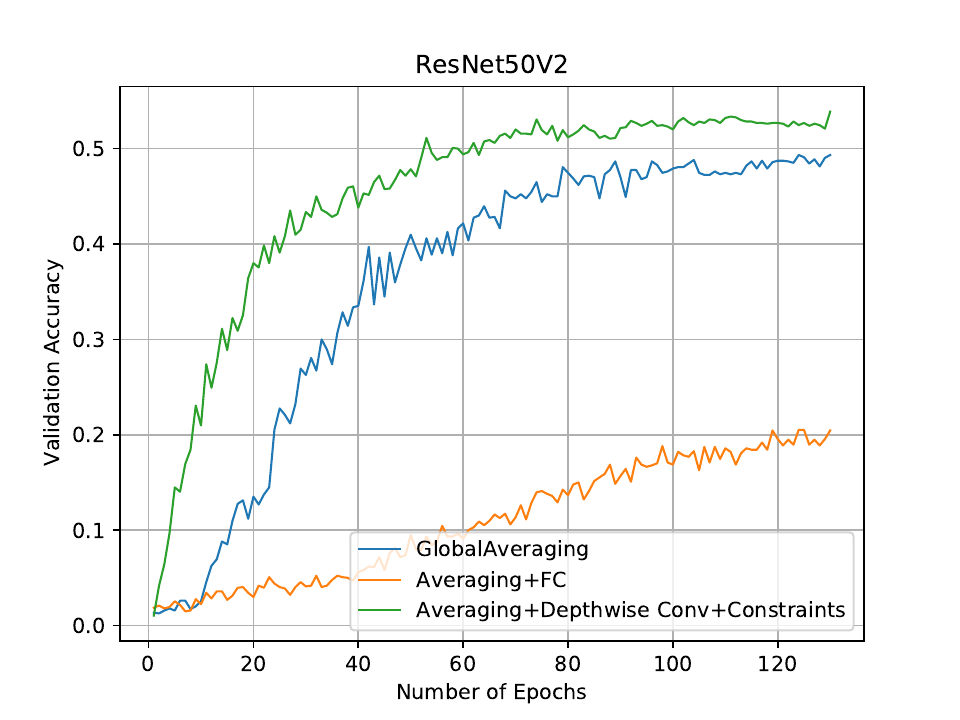}}
\newline
\subfloat[Xception training accuracy per epochs]{\label{512-mit-acc-Xception}\includegraphics[width=0.5\linewidth]{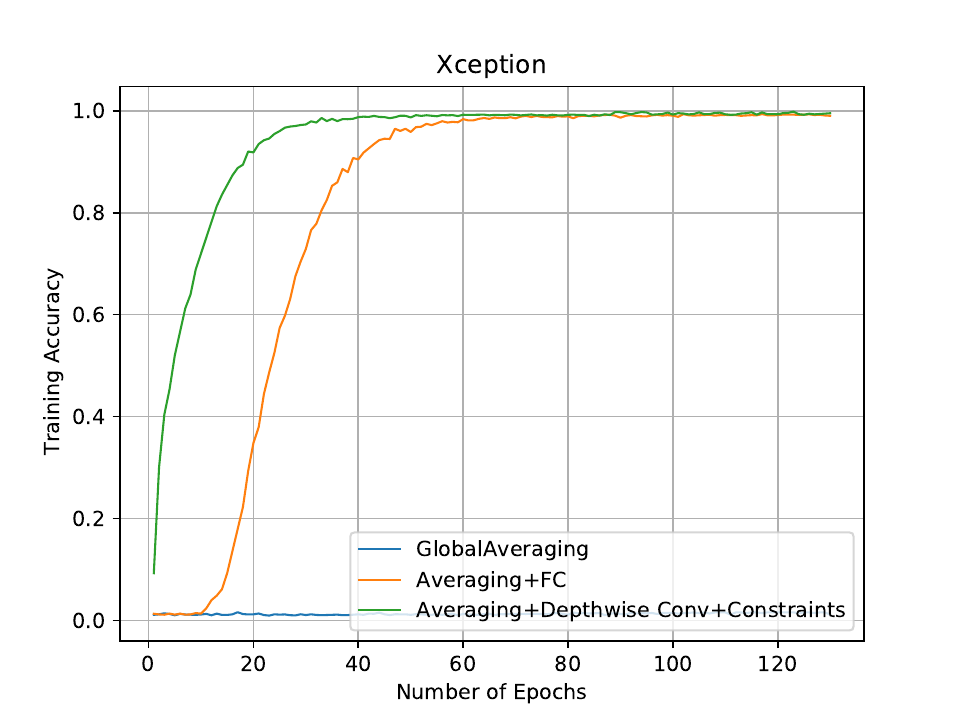}}
\subfloat[Xception validation accuracy per epochs]{\label{512-mit-val-Xception}\includegraphics[width=0.5\linewidth]{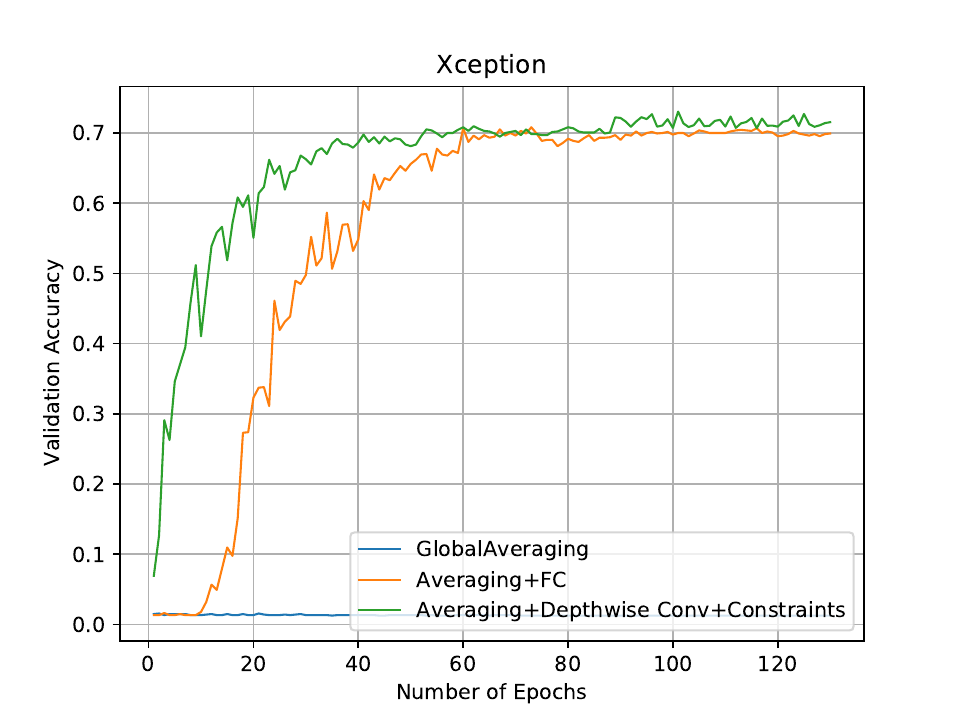}}

\caption{This figure displays the validation and training accuracies of the trained models with different classification layers on the MIT Indoors Scenes dataset for 130 epochs.  In this experiment, the images were resized to 512×512 pixels. The Depthwise convolutional layer and the Averaging layer's kernel size were 5×5 and 3×3, respectively. }
\label{512-mit-all}
\end{figure}

Table. \ref{512-mit} and figure \ref{512-mit-all} show the obtained results on MIT Indoors Scenes Dataset.

\begin{table*}[!ht]
\centering
\large
\caption{This table displays all the parameters used for training the models on 512x512 images.}
\begin{tabular}{l|l|}
\cline{2-2}
                                                                                                                      & MIT Indoors Scene                                                                         \\ \hline
\multicolumn{1}{|l|}{Batch Size}                                                                                      & 15                                                                                        \\ \hline
\multicolumn{1}{|l|}{Image Size}                                                                                      & 512x512                                                                                   \\ \hline
\multicolumn{1}{|l|}{Optimizer}                                                                                       & SGD (momentum:0.9)                                                                        \\ \hline
\multicolumn{1}{|l|}{Initial Learning rate}                                                                           & 0.045                                                                                     \\ \hline
\multicolumn{1}{|l|}{Learning rate decay}                                                                             & 0.94 every 2 epochs                                                                       \\ \hline
\multicolumn{1}{|l|}{Epochs}                                                                                          & 130                                                                                       \\ \hline
\multicolumn{1}{|l|}{Transfer Learning}                                                                               & Yes (ImageNet Pre-trained Weights)                                                        \\ \hline
\multicolumn{1}{|l|}{\begin{tabular}[c]{@{}l@{}}Depthwise Convolution Layer\\ Kernel Weight Initializer\end{tabular}} & \begin{tabular}[c]{@{}l@{}}Random Normal Initialization\\ (Mean:0, Std:0.01)\end{tabular} \\ \hline
\multicolumn{1}{|l|}{\begin{tabular}[c]{@{}l@{}}Depthwise Convolution Layer\\ Bias Initializer\end{tabular}}          & Zero Bias Initializtion                                                                   \\ \hline
\end{tabular}

\label{512-param}
\end{table*}

\begin{table*}[!ht]
\centering
\large
\caption{This table presents the results of our experiments on 512x512 images of the MIT Indoors Scenes dataset. For each deep convolutional model, three classification architecture has been studed. Model+GAP, Model+Avg+Dense, and Model+Avg+Depthwise Conv+Constraints are fully expressed in figures \ref{512-gap}, \ref{512-avg-dense}, and \ref{512-avg-depthwise-const}, respectively. Parameters show the number of weights of each model.}
\begin{adjustbox}{width=1\linewidth}
\begin{tabular}{|l|l|l|l|l|}
\hline
Model name  & Model + Classification type                & Parameters & Top-1 accuracy  & Top-5 accuracy  \\ \hline
            & Xception+GAP                               & 20,998,763 & 0.0157          & 0.0754          \\ \cline{2-5} 
Xception    & Xception+Avg+Dense                         & 24,291,947 & 0.7041          & 0.9106          \\ \cline{2-5} 
            & Xception+Avg+Depthwise Conv+Constraints    & 21,052,011 & \textbf{0.7362} & \textbf{0.9457} \\ \hline
            & ResNet50V2+GAP                             & 23,702,083 & 0.4933          & 0.7813          \\ \cline{2-5} 
ResNet50V2  & ResNet50V2+Avg+Dense                       & 26,995,267 & 0.2052          & 0.503           \\ \cline{2-5} 
            & ResNet50V2+Avg+Depthwise Conv+Constraints  & 23,755,331 & \textbf{0.5388} & \textbf{0.8149} \\ \hline
            & DenseNet201+GAP                            & 18,450,691 & 0.3433          & 0.6313          \\ \cline{2-5} 
DenseNet201 & DenseNet201+Avg+Dense                      & 21,538,051 & 0.1149          & 0.3321          \\ \cline{2-5} 
            & DenseNet201+Avg+Depthwise Conv+Constraints & 18,500,611 & \textbf{0.6037} & \textbf{0.8597} \\ \hline
\end{tabular}
\end{adjustbox}

\label{512-mit}
\end{table*}





The left plots in figure \ref{512-mit-all} present the training accuracy growth of each model. Based on these figures, the results are mixed and interesting. Xception is failed to be trained on the GAP layer. This means that the GAP layer (traditional way for classification) can not be applied to the Xception model when the images are large, and the number of classes is not few (we had 67 classes). By looking at the validation accuracy of Xception in figure \ref{512-mit-all} and Table. \ref{512-mit}, we can see this model with averaging and fully connected (FC) layer (figure \ref{512-avg-dense}) will be able to be trained and learn the images features. However, still, our proposed architecture obtains 3.21\% more Top-1 accuracy while containing 3 million fewer parameters. While both our architecture and averaging with FC layer are capable of learning spatial resolution, as our architecture is more optimized and has fewer parameters, it will face overfitting less than the other model and show better results.

Although Xception with Averaging and FC  layer can learn image features, this architecture fails on ResNet50V2 and DenseNet201 due to the high underfitting. This performance means that as there are many parameters in the final classification layer, it makes the models unable to learn and cause underfitting. figures \ref{512-mit-acc-DenseNet201} and \ref{512-mit-acc-ResNet50V2} prove this statement.

DenseNet201 can be trained with the GAP layer poorly (figure \ref{512-mit-acc-DenseNet201}), and we can consider that the GAP layer does not work well on this model, too. DenseNet201 with our architecture achieves near 26 \% more Top-1 accuracy than this model with the GAP layer.

ResNet50V2 training plot (figure \ref{512-mit-acc-ResNet50V2}) shows that this model can be trained properly with the GAP layer, but our architecture improves the Top-1 accuracy by 4.55 \% while having the same computational cost (Table. \ref{512-mit}).

Based on these results, the GAP layer and the averaging with FC architectures are not trustworthy and reliable when the images' resolution is high, and the number of classes is not few. Both of these architectures may work on some models and fail on other models. It means that in these situations (having large images and several classes), our proposed architecture is not just for improving the results, and it may be the only solution for training the models.
We can also conclude that one reason the developers did not use larger images on the ImageNet dataset (which includes 1000 classes) was that implementing deep models with the GAP layer fails on learning, and they could not reach acceptable results.

\subsubsection{Feature Visualization}
\label{321}

\begin{figure}[!ht]
\centering
\subfloat[Plotted features produced by models utilizing the GAP layer]{\label{grad-normal}\includegraphics[width=\linewidth]{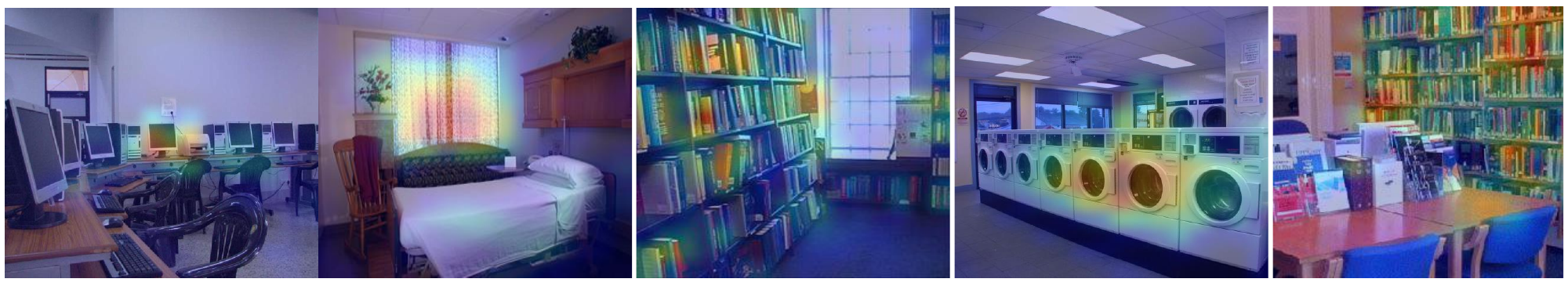}}
\newline
\subfloat[Plotted features produced by models utilizing our proposed architecture]{\label{grad-ours}\includegraphics[width=\linewidth]{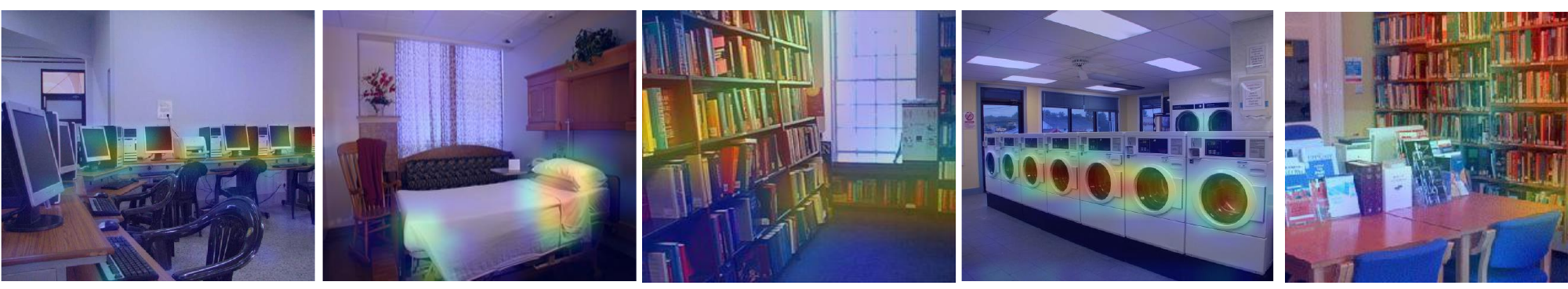}}
\caption{By looking at this figure, you can compare the quality of extracted features using the GAP layer and our architecture. The Grad-Cam algorithm has produced the features heatmaps. All the images in this figure have been classified correctly.}
\label{grad}
\end{figure}

To better show the performance of our architecture on convolutional models, we have adopted the Grad-CAM algorithm \cite{Selvaraju_2019} to highlight the feature maps on the images. We have applied this algorithm on some of the validation images of the MIT Indoors Scenes dataset that were classified by ResNet50V2 or DenseNet201 models. For each image, once we run the model with our architecture and another time, we executed the same model with the GAP layer on that image to compare the operation of these two methods.
figure \ref{grad}. represents the plotted feature heatmaps on the input images. 

It must be mentioned that the analyzed images are all correctly classified by both our architecture and the GAP. Our goal of this comparison is to show that even in the correctly classified images, the extracted features heatmaps confirm that models with our architecture are more robust in extracting spatial data of images.

\subsection{Methods that didn't work}
\label{33}

In section \ref{224}, we mentioned that one effective way to limit overfitting is the usage of an average pooling layer before the depthwise convolutional layer for reducing the kernel size of the feature map. This method is more practical when the input images are larger. We have tried several other methods besides the average pooling layer, but we observed that this layer performed better than all of them. Here we wish to express some of the other implemented methods which did not work:

\begin{enumerate}
\item Applying two depthwise convolutional layers instead of one. In this manner, the first depthwise convolutional layer will be used instead of the average pooling layer, and the second one extracts the final compressed features. As there are two depthwise convolutional layers, the kernel size of each one becomes smaller; therefore, the overfitting possibility decreases. Although this method seems more effective, our experiments determined this architecture causes underfitting, and the models could not be trained.
\item Utilizing interpolation algorithms for reducing the size of the feature maps. This idea was based on the assumption that using interpolation algorithms for resizing feature maps could keep the information more original than average pooling layers. Bilinear, bicubic, nearest, and area interpolations were studied and tested. Our investigations showed that using these interpolation algorithms will not result in higher accuracy than the average pooling.
\item Adding padding to the average pooling layer. As averaging with zero or unit padding changes the actual value of border features, it weakens the results.
\item Having strides on the average pooling layer. By applying stride, we could use smaller averaging kernels and increase the quality of the averaged feature map. But as averaging with stride combines the neighbor features, and the correlation between neighbor averaged features increases,  it does not work well, and the results downgrade.
\end{enumerate}

Based on these experiments, we found that adding a small average pooling layer before the depthwise convolutional layer will obtain better performance.

\section{Discussion}
\label{4}

This research focuses on solving the problem of losing spatial resolution of images caused by the Global Average Pooling (GAP) layers.
We came up with this idea to replace the GAP layer with a new set of layers which let the model learn how to weighted-sum the values of feature map with the trained weights and squeeze it into a compressed array wisely.  
Our final proposed architecture, called Wise-SrNet is decribed in sections. \ref{223} and \ref{224} and is depicted in figure \ref{224-avg-depthwise-const-dp}.

Based on the input images' resolution, our analyses were divided into two categories : 224×224 and 512,×512. We explored three image classification datasets for investigating our models: a part of the ImageNet dataset, MIT Indoors Scenes, and Intel Image Classification datasets. Our architecture was implemented on various models of the three famous convolutional families of ResNet, DenseNet, and Inception families. DenseNet121, DenseNet169, DenseNet201, ResNet50, ResNet50V2 and Xception models were utilized in our research.

Upon the 224×224 resolution images, our architecture showed 3\%-6\%,
 2\%-6\%, and 5\%-8\% higher Top-1 accuracy on the sub-ImageNet, Intel Image Classification, and MIT Indoors Scenes datasets, compared to the GAP layer, respectively.
Among all of the evaluated models, ResNet50 is more adaptable with our methods, and our architecture dispenses better improvement on this model than other models. The ResNet50 model with our architecture showed near 5-8\% growth in validation accuracy.

Considering the 512×512 resolution images, we ran our investigations on the MIT Indoors Scenes dataset.
ResNet50V2, DenseNet201, and Xception models were chosen for running at this stage of our work. Three classification versions for each model were studied: Our architecture, the GAP layer, and the classic classification model based on flattening and FC layers (figure \ref{512-avg-dense}).

After running tests on 512×512 resolution images, we observed that both the GAP layer and the classic method could not be trusted when there are several classes of images or the image are in high resolution. Both of these architectures failed on some models in the mentioned circumstance. The GAP layer did not work on Xception and worked very weak on DenseNet201, and the classic method did not work on ResNet50V2 and DenseNet201. Still, our architecture performed very well on all the models and achieved acceptable results. In this condition, our architecture may not be just for improving the accuracy, but it can be the main solution for obtaining reliable results.
Our architecture improved the Top-1 accuracy by 3-\% to 26\% on the MIT Indoors Scenes dataset, including 67 classes of images with 512×512 resolution.

Our proposed technique can be applied to many networks and can create more accurate models on the ImgaNet or other popular benchmarks. As there are 1000 classes in the ImageNet dataset, increasing the images' resolution may be impossible for the developers because the GAP layer could fail the models to learn in this condition. Hence, our architecture showed to be adaptable and still achieved better accuracy under challenging situations like having lots of classes and large images.
Accordingly, using our methods, the developers can also increase the images' resolution on the ImageNet dataset and reach much better classification accuracy.

Now, we wish to explain one of our important observations. Although we reported the results of six deep convolution models in this paper, we also had evaluated our results on many families of convolutional models. Our architecture showed improvement in most cases, as expected, but for the EfficientNet family \cite{tan2020efficientnet}, we did not witness remarkable progress in performance. Hence, the logic and observations are based on the fact that analyzing the spatial resolution with our architecture must enhance classification performance, but why doesn't this fact apply to the EfficienNet family?

The main reason behind this matter can be summarized as follow.
 The EfficientNet architectures are not hand-designed and have been developed by Neural Architecture Search (NAS) engine \cite{zoph2017neural}.
These models' components are inspired by the MobileNet architecture \cite{howard2017mobilenets}, and the depth, width, and resolution of their layers have been scaled for best performance. The essential point is that the NAS had designed the EfficienNet architectures and scaled the layers' parameters when the Global Average Pooling layer was placed at the end of the models for features compression. Based on this reason,  we can conclude that the EfficientNet models have been produced by a neural search engine to acquire the best performance on the GAP layer, which has affected the type of their architectures and layers scaling parameters. 

 This is the main reason why EfficientNet did not achieve the same improvement with our architecture as other models since it is fine-tuned for working with the GAP layer. For feature work, other researchers can use the same approach of EfficientNet, but this time replace the GAP layer with our classification layers to create the next stage of the Efficientnet models, which are more robust and accurate than the existing models.

\section{Conclusion}
\label{5}

Global Average Pooling (GAP) layer usage is currently the most common method for compressing the feature maps before the classification layers. After the VGG models, most of the following deep convolutional models utilized the GAP layer in their architecture for diminishing computational costs.
Although the GAP layer minimizes the number of model weights, its disadvantage is removing the spatial resolution of feature maps.
This paper solved this problem by introducing Wise-SrNet, a novel architecture that processes the feature maps' spatial data without increasing computation costs. 

Our proposed method allows the model to create an independent equation for each channel of the feature map through various sets of trainable weights for analyzing the spatial values.
Unlike many other works, our architecture is not limited to any function like averaging and does not apply the same weights for all the feature map's channels.
As our architecture can process both of the spatial and channel data and carry almost the same computational cost as the GAP layer, it can be the best replacement for this classic classification technique.

The central core of Wise-SrNet is made of our proposed depthwise convolutional layer with some specific parameters that help the model analyze all the feature maps' data. But this layer can not work alone because it makes the model overfitted.
So, we have also added some other layers and constraints to the depthwise convolutional layer to improve its learning ability and decrease overfitting.

In this research, three datasets with different resolutions were investigated for our experiments, which include a part of the ImageNet dataset, MIT Indoodrs scenes, and the Intel Image classification challenge. DenseNet121, DenseNet169, DenseNet201, ResNet50, ResNet50V2, and Xception models were also studied and compared on ours and other classification architectures, including the GAP layer. 
Our architecture improved the Top-1 accuracy between 2\% to 8\% for images with 224×224 resolution and 3\% to 26\% for images with 512×512 resolution, respectively.

Our observations confirm that adopting the classic methods for compressing the feature maps and classification; is not reliable on datasets with many classes and large images. As our novel architecture is optimized and performs accurately in these challenging circumstances, it can greatly improve image classification benchmarks.

\section{Code Availability} 
\label{6}
All the main codes of our paper have been shared online for public use at \href{https://github.com/mr7495/image-classification-spatial}{https://github.com/mr7495/image-classification-spatial}

\bibliographystyle{abbrv}  
\bibliography{arxiv}  

\end{document}